\documentclass[journal]{IEEEtran}

\usepackage{times}
\usepackage{epsfig}
\usepackage{graphicx}
\usepackage{amsmath}
\usepackage{amssymb}
\usepackage[commentsnumbered]{algorithm2e}
\usepackage{threeparttable}
\usepackage{booktabs, dcolumn}
\usepackage{stfloats}
\usepackage{adjustbox}
\usepackage{array}
\usepackage{booktabs}
\usepackage{multirow}
\usepackage{rotating}
\usepackage{color}
\usepackage{subcaption}
\usepackage{float}
\usepackage{colortbl}

\definecolor{LightCyan}{rgb}{0.88,1,1}

\begin{document}
%
\title{Scene Parsing with Integration of Parametric and Non-parametric Models}
%
%
%

\author{Bing~Shuai,~\IEEEmembership{Student Member,~IEEE,}
        Zhen~Zuo,~\IEEEmembership{Student Member,~IEEE,}
        Gang~Wang,~\IEEEmembership{Member,~IEEE}
        ~and~Bing~Wang,~\IEEEmembership{Student Member,~IEEE}
\thanks{B.Shuai, Z.Zuo, G.Wang and B.Wang are with the School
of Electrical and Electronic Engineering, Nanyang Technological University, Singapore.
E-mail: \{bshuai001, zzuo1, wanggang, wang0775\}@ntu.edu.sg.}
\thanks{Correspondence should be directed to G. Wang.}}

\maketitle

\begin{abstract}
  We adopt Convolutional Neural Networks (CNNs) to be our parametric model to learn discriminative features and classifiers for local patch classification. Based on the occurrence frequency distribution of classes, an ensemble of CNNs (CNN-Ensemble) are learned, in which each CNN component focuses on learning different and complementary visual patterns. The local beliefs of pixels are output by CNN-Ensemble.
  Considering that visually similar pixels are indistinguishable under local context, we leverage the global scene semantics to alleviate the local ambiguity. The global scene constraint is mathematically achieved by adding a global energy term to the labeling energy function, and it is practically estimated in a non-parametric framework.
  A large margin based CNN metric learning method is also proposed for better global belief estimation.
  In the end, the integration of local and global beliefs gives rise to the class likelihood of pixels, based on which maximum marginal inference is performed to generate the label prediction maps.
  Even without any post-processing, we achieve state-of-the-art results on the challenging SiftFlow and Barcelona benchmarks.
\end{abstract}


%
\IEEEpeerreviewmaketitle

\section{Introduction}
\label{Section:Intro}

\IEEEPARstart{S}{cene} parsing (also termed as scene labeling, scene semantic segmentation) builds a bridge towards deeper scene understanding. The goal is to associate each pixel with one semantic class. Generally, ``thing" pixels (car, person, etc) in real world images can be quite visually different due to their scale, illumination and pose variation, meanwhile ``stuff" pixels are usually visualy similar (road, sea, etc) in a local close-up view. Hence, the local classification for pixels is challenging. Besides, the class frequency distribution is highly imbalanced in natural scene images: more than 80\% pixels in the images belong to only a few number of semantic classes. Thus, the classification model is biased towards frequent classes due to the scarcity of training instances for rare classes. Overall, these issues pose scene parsing as one of the most challenging problems in computer vision.

\begin{figure}[t]
\centering
\includegraphics[width=0.365\textwidth]{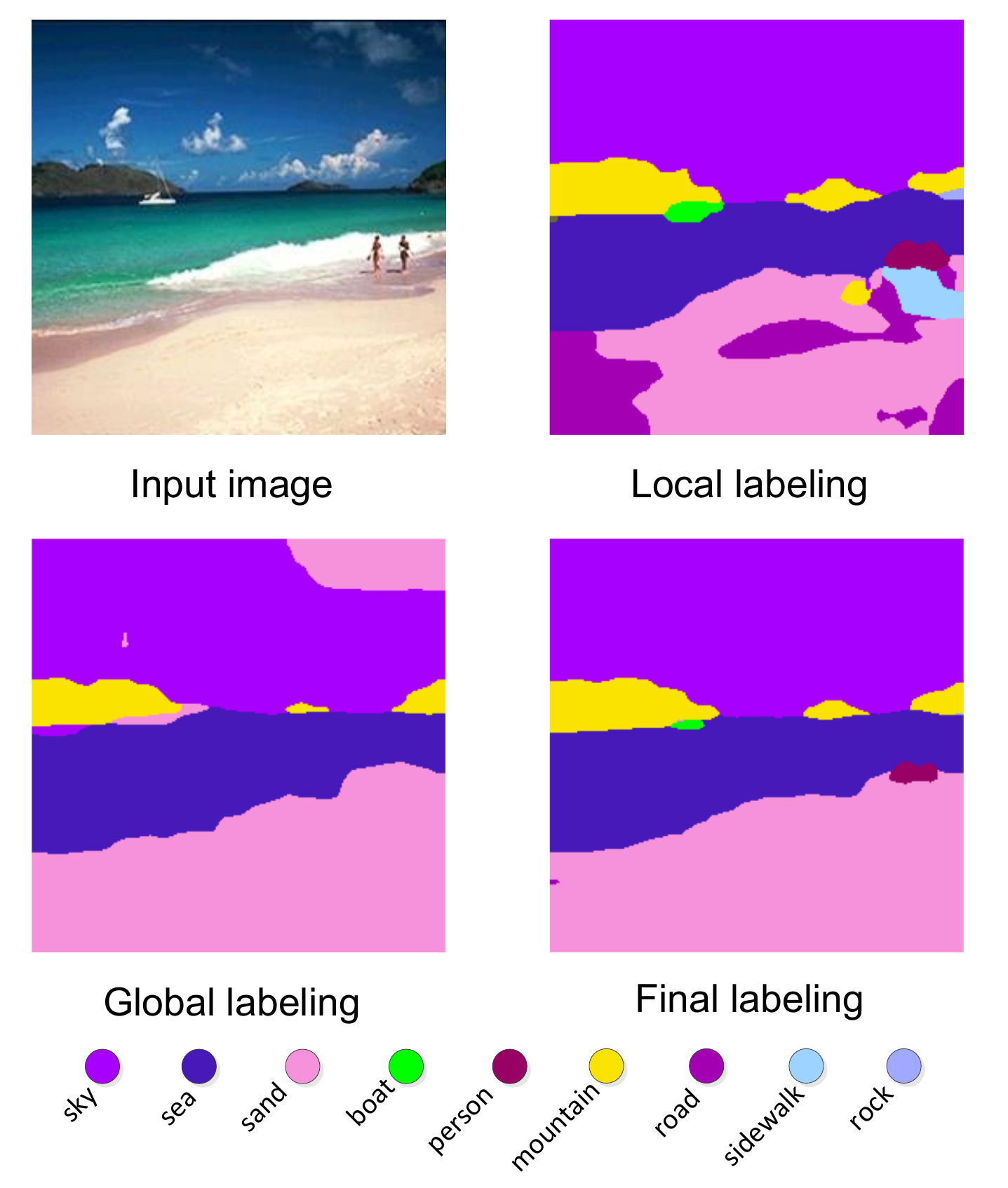}
\caption{Motivation of our integration model: the parametric model can distinguish visually different pixels very well, but get confused for pixels that are visually similar in local context. However, the local ambiguities can be easily eliminated as long as the correct global scene semantics is revealed. A more consistent labeling result can be achieved by integrating their beliefs. The figure is best viewed in color.
}
\label{Fig:intro}
\end{figure}

The recent advance of Convolutional Neural Networks (CNNs) \cite{krizhevsky2012imagenet,lecun1998gradient} has revolutionized the computer vision community due to their outstanding performance in a wide variety of tasks \cite{krizhevsky2012imagenet,oquab2013learning,wolfdeepface,zhang2013panda,tompson2014joint,zeiler2014visualizing}. Recently, Farabet et al. \cite{farabet2013learning} and Pinheiro et al. \cite{pinheiro2014recurrent} has applied CNNs to scene labeling.  In this scenario, CNNs are used to model the class likelihood of pixels directly from local image patches. They are able to learn strong features and classifiers to discriminate the local visual subtleties. In general, single CNN fails to produce satisfactory labeling results due to the severely imbalanced class frequency distribution in natural scene images (as exampled in Figure \ref{Fig:siftFlow-sampling}). \textcolor{black}{To address this issue, we propose the \textbf{CNN-Ensemble}}, which aggregates the predictions from different and complementary CNNs. The CNN component shares the identical network architecture, but it is trained from image patches with disparate statistics, which are generated from different sampling methods. In the end, the proposed CNN-Ensemble is capable of yielding more reliable labeling results than any single CNN.

Even though powerful, CNN-Ensemble still struggles in differentiating visually similar pixels as it only consider limited context in local classification.  As exampled in Figure \ref{Fig:intro}, the sand pixels are confused with road and sidewalk pixels in a local view. We refer to such problem as \textit{local ambiguity}. This problem has usually been addressed from two perspectives:
 \begin{itemize}
  \item Augmenting the scale of context to represent pixels: \cite{farabet2013learning} considers multi-scale context input, \cite{pinheiro2014recurrent} increases the size of context input in a recurrent CNN framework. These methods somehow mitigate the local ambiguity, however they may have an negative effect to small-size objects and may also degrade the efficiency of the system.
  \item Building a probabilistic graphical model to capture the explicit label dependencies among pixels \cite{he2004multiscale,kohli2009robust,shotton2006textonboost,zhang2012efficient}. However, the parametric graphical model is usually hard and inefficient to optimize when the higher order potentials are involved, and the low-order potentials suffer from low representation power.
 \end{itemize}
Here in this paper, we propose to utilize the global scene semantics to eliminate the \textit{local ambiguity}. As a simple example in Figure \ref{Fig:intro}, the confusion between `road' and `sand' pixels can be easily removed if the global ``coast" scene is revealed.  Intuitively, a global scene constraint is implicitly enforced to allow more reliable local classification. Such global constraint is mathematically achieved by adding a global energy term to the labeling energy function. However, due to the extraordinarily huge labeling space, it's infeasible to model the global energy parametrically. Thus, the global energy is practically modeled in a non-parametric framework by transferring the class dependencies and priors from its global similar exemplar images.



Furthermore, a large margin based metric learning objective is introduced to fine tune the network, thus making the estimation of global belief more accurate. Finally, the class likelihood of pixels are obtained by integrating the local and global beliefs. Based on which, our integration model outputs the label prediction maps. We justify our method on the popular and challenging scene parsing benchmarks: SiftFlow \cite{liu2009nonparametric} and Barcelona \cite{tighe2010superparsing} datasets. Even without any post-processing, our integration model is able to achieve very competitive results that are on par with the state-of-the-arts. Overall, the contributions of this paper are summarized as follows:
\begin{enumerate}
  \item We propose the \textbf{CNN-Ensemble}, in which each CNN component concentrates on learning distinct visual patterns. The aggregation of single CNNs gives rise to much more powerful model that is able to generate more reliable labeling maps.
  \item We leverage global scene semantics to remove the \textit{local ambiguity by} transferring class dependencies and priors from similar exemplars.
  \item We introduce the CNN metric, and show that the learned metrics are beneficial in our non-parametric global belief estimation.
\end{enumerate}

This paper is an extension to the conference paper \cite{shuai2015integrating}. The rest of the paper is organized as follows: section \ref{Section:Related} firstly reviews, discusses and compares our methods with relevant works. Following that, the formulation of our integration model are presented in Section \ref{Section:Motivation}; Then, details of estimating the local and global beliefs are elaborated in Section \ref{Section:local_belief} and \ref{Section:global_belief} respectively; Section \ref{Section:Experiment} demonstrates the experimental setup and reports the results of the proposed methods; Section \ref{Section:Conclusion} concludes the paper.


\section{Related Work}
\label{Section:Related}
Scene parsing has attracted more and more attention in recent years. Among all the interesting works, we review and discuss four line of works that are most relevant to ours.

\subsection{Feature Learning}
The first direction exploits extracting better features for classifying pixels/superpixels. Previously, low-level and mid-level hand-crafted features are designed to capture different image statistics. They usually lack discriminative power and suffer from high dimensionality, thus limiting the complexity of the full labeling system. Recently, machine learning techniques are commonly used to learn discriminative features for various computer vision tasks \cite{farabet2013learning,pinheiro2014recurrent,wang2015video,zuo2015exemplar}. In accordance, Farabet et al. \cite{farabet2013learning} fed a convolutional neural network with multi-scale raw image patches, and they have presented very interesting results on real-world image scene labeling benchmarks. Furthermore, Pinheiro et al. \cite{pinheiro2014recurrent} adopted a recurrent CNN to process the large-size image patches. Bulo et al. \cite{buloneural} learned a more compact random forest by substituting the random split function with a stronger Neural Network.
\textcolor{black}{Mostajabi et al. \cite{mostajabi2015robust}\cite{mostajabi2015feedforward} extracted features from different scopes of image regions, and then concatenated them to yield the context-aware representation. Therefore, regional and global context is encoded in the local representation.}
In their works, the local disambiguation is achieved via augmenting input context. In contrast, we leverage the global scene constraint to mitigate the local ambiguity.

\subsection{Probabilistic Graphical Models}
Another line of works focus on exploring explicit dependency modeling among labels, which is usually formulated as a structure learning problem. Shotton et al. \cite{shotton2006textonboost} formulated the unary and pairwise features in a 2nd-order sparse Conditional Random Fields (CRF) graphical model. Roy et al. \cite{royscene}, Zhang et al. \cite{zhang2012efficient} and Chen et al. \cite{chen2014semantic} built a fully connected graph to enforce higher order labeling coherence. Kohli et al. \cite{kohli2009robust}, Kontschieder et al. \cite{kontschieder2011structured} and Marquez et al. \cite{marquez2014non} modeled the higher order relations by considering patch/superpixel as a clique.  He et al. \cite{he2004multiscale} defined a multi-scale CRF that captures different contextual relationships ranging from local to global granularity. \textcolor{black}{Zheng et al. \cite{zheng2015conditional} formulates the CRF as a neural network, so its inference equals to applying the same neural network recurrently until some fixed point (convergence) is reached.}
Recently, Shuai et al. \cite{shuai2015quaddirectional,shuai2015dag} adopt recurrent neural networks (RNNs) to propagate local contextual information and it shows superiority over PGMs on the applicability to large-scale scene parsing task.
Our work is related to this branch of works, but approaches from a different angle. The potentials in these works are usually modeled parametrically, therefore extensive efforts are needed for learning these parameters. Our global energy term can be estimated very efficiently in a non-parametric framework.

\subsection{Label Transfer Models}
Recently, non-parametric label transfer methods \cite{liu2009nonparametric,tighe2010superparsing,eigen2012nonparametric,singh2013nonparametric,tung2014collageparsing,yangcontext} have gained popularity due to their outstanding performance and the scalability to large scale data. They usually estimate the class likelihood of image unit from the globally similar images. In a nutshell, global scene features are firstly utilized to retrieve the relevant images, whose label maps are then leveraged to estimate the class likelihood of image units.
The pioneering label transfer work \cite{liu2009nonparametric} transformed RGB image to SIFT \cite{lowe2004distinctive} image, which was used to seek correspondences over pixels. Then, an energy function was defined over pixel correspondences, and the label prediction maps are obtained by minimizing the energy. The Superparsing system \cite{tighe2010superparsing} performed label transfer over image superpixels.
Eigen et al. \cite{eigen2012nonparametric} learned adaptive weights for each low-level features, and it resulted in better nearest neighbor search.
Gould et al. \cite{gould2012patchmatchgraph,gould2014superpixel} built a graph for dense image patch and superpixel to achieve the label transfer. We adopt this framework to evaluate our global energy term. In comparison with these works that are based on hand-crafted features, we used the learned CNN features which are more compact and discriminative. We expect our features to benefit their systems in terms of accuracy and efficiency as well.


\subsection{Ensemble Models}
The ensemble methods \cite{zhou2012ensemble}\cite{shotton2013decision} have achieved great success in machine learning. The idea is to build a strong predictors by assembling many weak predictors. The assumption is that these weak predictors are complementary to each other when they are trained with different subset of features and training data. Some examples are random forest \cite{breiman2001random}\cite{criminisi2013decision}, bagging \cite{breiman1996bagging}, boosting \cite{freund1999short}, etc. Random forest has been successfully used in solving image labeling problems. For example, Shotton et al. \cite{shotton2013decision} learned an ensemble of decision jungles to output the semantic class labels of pixels.  Kontschieder et al. \cite{kontschieder2011structured} constructed a random forest that directly maps the image patches to their corresponding structured output maps. Our CNN-Ensemble is different from these works. First, the individual model in theirs are very weak, whereas each of our single CNN has very strong capability. Second, the data that are used for each individual model training in their works are sampled without discrimination. In contrast, data that are fed to each single CNN are sampled differently, therefore their statistics are different.

Recently, the ensemble models \cite{krizhevsky2012imagenet}\cite{simonyan2014very}\cite{szegedy2014going} have been pervasively adopted in the large-scale ImageNet classification competition \cite{russakovsky2014imagenet}. Specifically, many deep neural networks (their network architeture is identical) are trained, and the fusion of their decisions give rise to the output. By doing this, the ensemble model is able to enhance its classification accuracy slightly. Our CNN-Ensemble also differs from these works.  In their works, each network component is trained with exactly the same data, and the difference of network components mainly originates from nonidentical network initializations. In contrast, every CNN component is trained from entirely different data, which will guide the CNN component to focus on learning different but complementary visual patterns.

\section{Formulation}
\label{Section:Motivation}

\begin{figure*}[t]
\begin{center}
\includegraphics[width=1.0\textwidth]{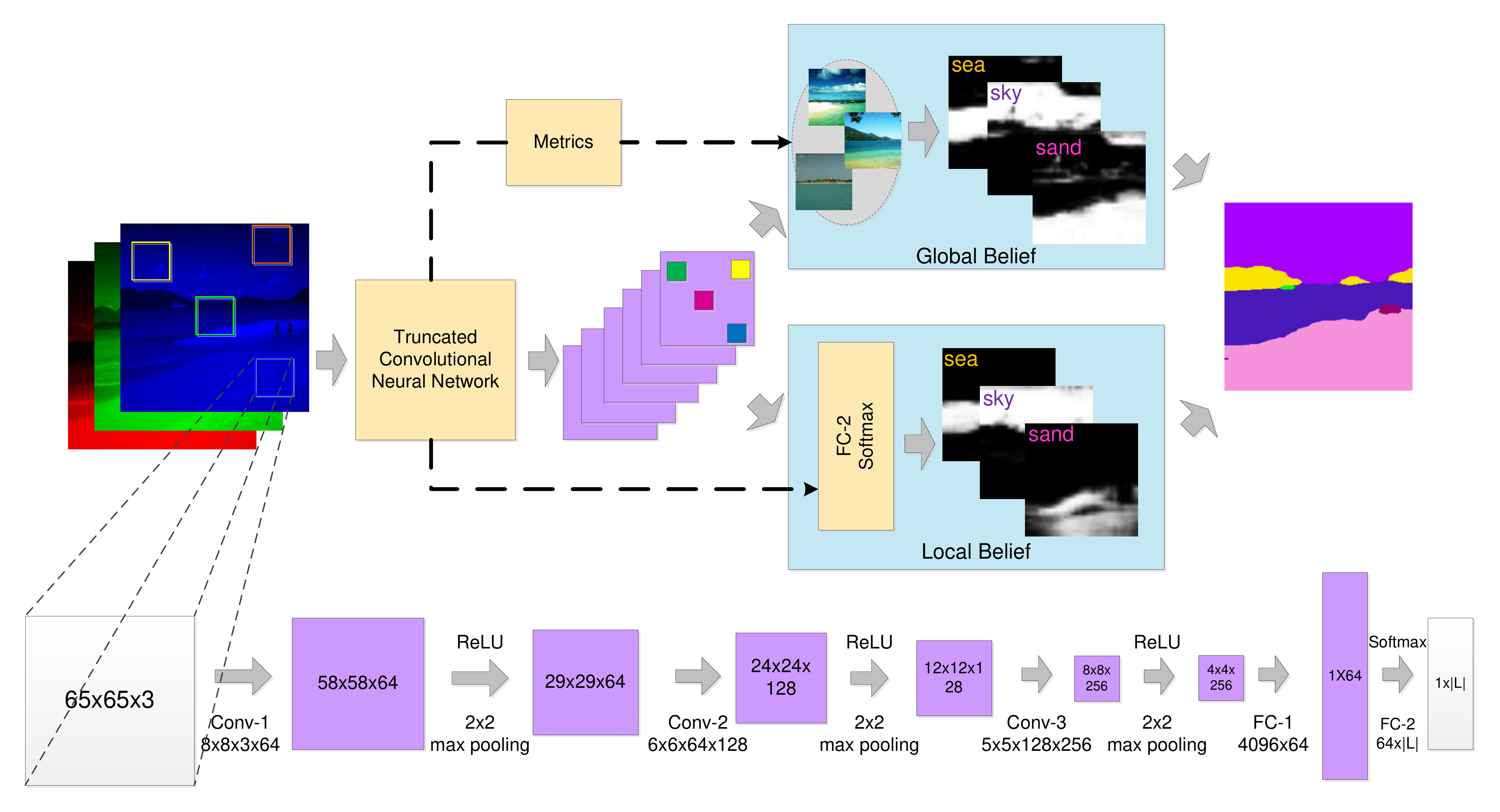}
\end{center}
\caption{Framework of our integration model: the parametric CNNs are responsible for emitting the local beliefs, and the non-parametric model is practically used to output the global beliefs. The integration of local and global beliefs gives rise to the un-normalized class likelihood of pixels. The modules painted in yellow represents parametric models (CNN and Metrics).}
\label{Fig:cnn}
\end{figure*}

The image labeling task is usually formulated as a discrete energy minimization problem. Specifically in this paper, we consider minimizing the following energy:
\begin{equation}
E(X,Y) = E_I(X,Y) + E_G(X,Y)
\end{equation}
where $X = \{X_1, X_2, \ldots, X_N\}$ is the observed image and $X_i$ corresponds to the $i$th pixel;  $Y = \{Y_1, Y_2, \ldots, Y_N\},\  Y \in \{1,2,\ldots |L|\}^N$ denotes a labeling configuration for image $X$; \textcolor{black}{$E_I(X,Y)$ and $E_G(X,Y)$ are the local and global energy term respectively.}

Individually speaking, the local energy term $E_I(X, Y)$ measures the likelihood of image $X$ taking the labeling configuration $Y$ in the local view. Mathematically, it is expressed as the summation of local unary potential $\Psi_I(X_i, Y_j)$:
\begin{equation}
E_I(X,Y) = \sum_{X_i\in X} \Psi_I(X_i, Y_j)
\end{equation}
where $\Psi_I(X_i, Y_j) = -log(P_I(X_i, Y_j))$ is defined as the negative log likelihood of pixel $X_i$ being labeled as $Y_j$;
Hereafter, we call $P_I(X_i, Y_j)$ the local belief.

\textcolor{black}{The global energy $E_G(X, Y)$, on the other hand, reflects the likelihood of image $X$ taking the labeling configuration $Y$ in the global view. A naive implementation is to consider all the pixels to be in a single clique, and $E_I(X,Y)$ evaluates the energy term according to the labeling states. However in this scenario, the labeling states are prohibitively huge ($|L|^N$), it makes the energy evaluation intractable in practice. Inspired by \cite{marquez2014non}, we adopt an non-parametric approach to decompose the global energy to the aggregation of global unary potential $\Psi_G(X_i, Y_j)$:
\begin{equation}
\label{Equation:global}
E_G(X,Y) = \sum_{X_i\in X} \Psi_G(X_i, Y_j)
\end{equation}
where $\Psi_G(X_i,Y_j) = -log(P_G^{\mathcal{S}(X)}(X_i,Y_j))$;  $P_G^{\mathcal{S}(X)}(X_i,Y_j))$ denotes the likelihood of $X_i$ being $Y_j$ in the global scene view, and it is estimated from images in $\mathcal{S}(X)$ which are expected to share similar global scene semantics with image $X$. Likewise, we call $P_G^{\mathcal{S}(X)}(X_i, Y_j)$ the global belief hereafter.}
As shown, the global energy implicitly captures the pixel dependencies via transferring scene semantics from images in $\mathcal{S}(X)$. As an example in Figure \ref{Fig:intro}, the global scene exemplars ($\mathcal{S}(X)$) define a "coast" scene, in which road and sidewalk pixels are invalid and sand pixels are more likely to appear in the bottom regions.
Finally, by rewriting the energy functions, we generate the following form:
\begin{equation}
E(X,Y) = -\sum_{X_i\in X}log((P_I(X_i,Y_j) \cdot P_G^{\mathcal{S}(X)}(X_i, Y_j)))
\label{Equation:main}
\end{equation}
The above energy  is numerically proportional to the integration of beliefs from two sources: (1), Local belief: $P_I(X_i,Y_j)$ measures the belief for pixel $X_i$ based on its surrounding local context; (2), Global belief: $P_G^{\mathcal{S}(X)}(X_i,Y_j)$ denotes the belief for $X_i$ from the global scene view. An intuitive interpretation of Equation \ref{Equation:main} is that a global scene constraint (prior) is enforced to the local classification in the form of weighting the local beliefs of pixels with their corresponding global beliefs. Since the estimation of class likelihood for different pixels is independent, the energy minimization (inference) can be done in an efficient pixel-wise manner: $Y = \bigcup_{i=1:N} Y_i, \ Y_i = argmin_{1,\ldots, |L|}E(X_i,Y_j)$.

The pipeline of our model is depicted in Figure \ref{Fig:cnn}. Systematically, an image is first passed to the truncated CNN, and the corresponding pixel feature maps are generated.
Next, the feature maps are fed into two branches: (1), they are independently classified based on the parametric CNNs (CNN-Ensemble), which yield the local beliefs; (2), they are aggregated to produce the global scene envelop, which is used to retrieve the global similar exemplar images. Based on which,  the global belief is estimated. Finally, the integration of local and global beliefs yields the un-normalized class likelihood of pixels, based upon which the integration model outputs the label prediction map. We elaborate each module in the following sections.

\section{Local Beliefs}
\label{Section:local_belief}

In a local view, the semantic class of each pixel $X_i$ is determined by its surrounding image region (patch). A parametric model is commonly used to emit its local belief $P_I(X_i, Y_j)$. In this paper, this model is parameterized by \textbf{CNN-Ensemble} - an ensemble of Convolutional Neural Network (CNNs). The CNN components are trained from entirely different image regions (patches) in terms of the class frequency distribution, thus they capture complementary image statistics. In detail,
each CNN component is enforced to focus on discriminating some specific classes by adaptively learning different features and classifiers. By fusing these complementary cues, CNN-Ensemble is expected to output more reliable labeling results.

\subsection{Parametric CNNs}
The Convolutional Neural Networks (CNNs) learn features and classifiers in an end-to-end trainable system. They are able to learn compact yet discriminative representations, which are easy to be differentiated for the jointly learned classifiers. Specifically in the image labeling task, the parameters of CNNs are optimized based on the image patches, whose labels are associated with the centering pixels. Unlike other CNNs \cite{farabet2013learning} \cite{pinheiro2014recurrent} which are fed with multi-scale or large field-of-view patches, we use moderate-size contextual windows to predict the labels for pixels. By doing this, we enforce the network to learn strong representations to discriminate the local visual subtleties. Meanwhile, the locality information is well preserved, which is crucial to differentiate small-size object classes.  Moreover, as evidenced by the experiments later, our CNNs outperform their nets \cite{farabet2013learning}\cite{pinheiro2014recurrent} dramatically. In addition, our CNNs are more efficient in terms of inference.

The architecture of our CNNs is demonstrated in Figure \ref{Fig:cnn}. It accepts $65 \times 65$ image patches as valid input. If we assume that the last fully connected layer (FC-2) serves as the functionality of classifiers, the removal of which in the CNN gives rise to the Truncated Convolutional Neural Network. In other word, if we pass an image patch ($65 \times 65$) to the truncated CNN, its output is a representation vector (64 dimension), which summarizes the contextual information of the input patch. In this perspective, truncated CNN can be interpreted as a feature extractor.


\subsection{Data Sampling}
\begin{figure}
  \centering
  \includegraphics[width=0.725\linewidth]{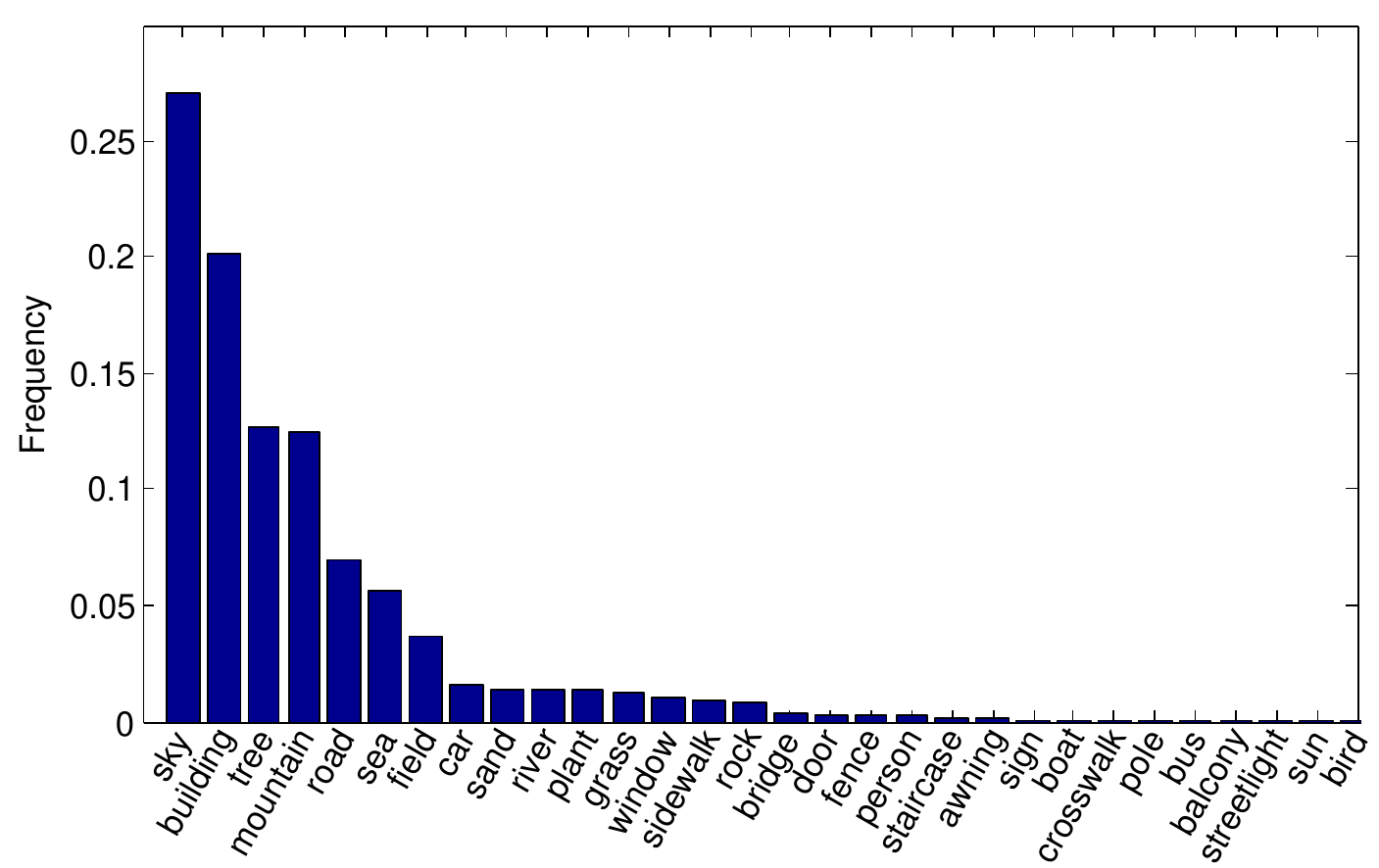}
  \includegraphics[width=0.725\linewidth]{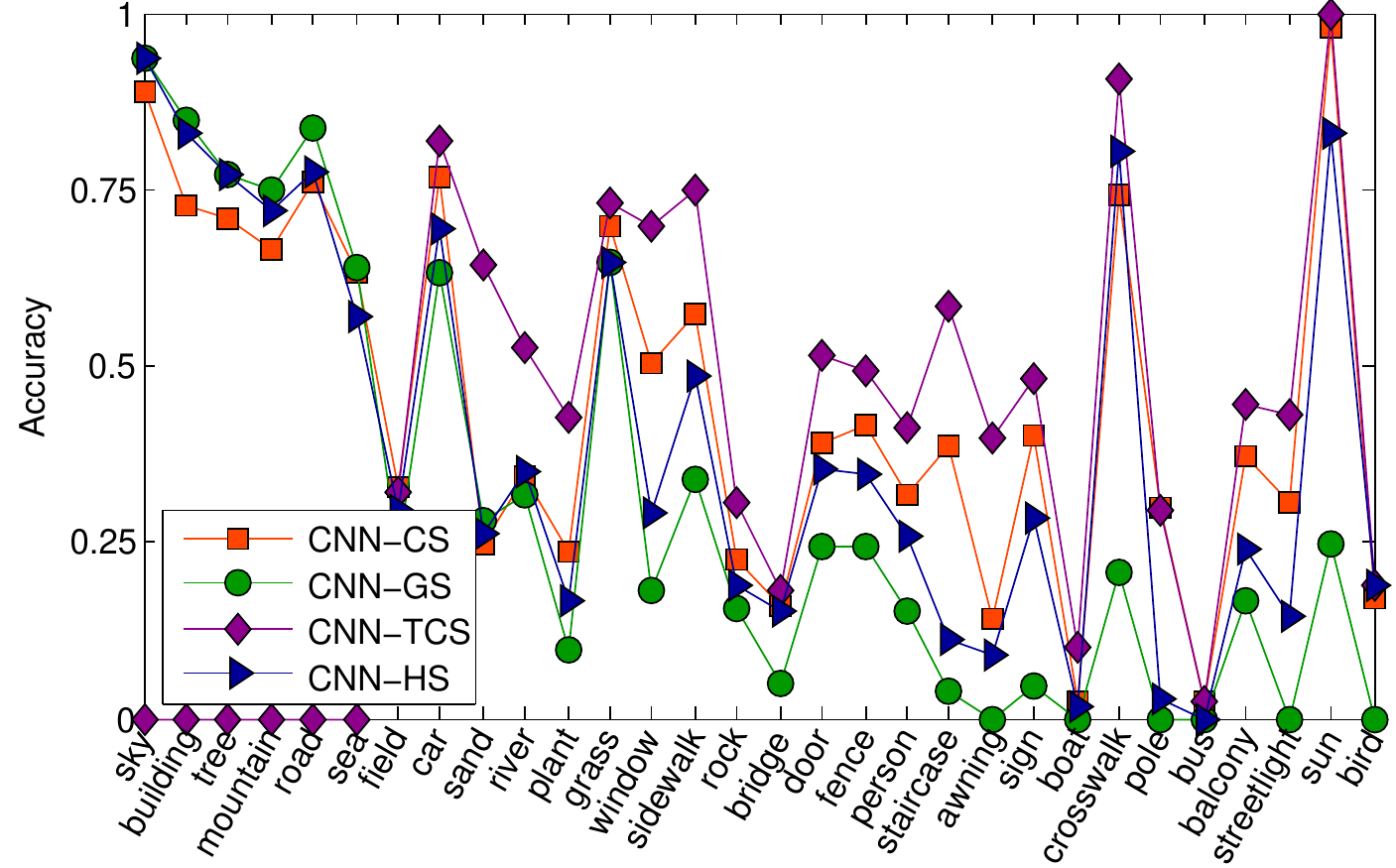}
  \caption{The first graphic shows an example of the class frequency distribution in natural scene images, through which we can observe that large percentage of pixels belong to very small number of frequent classes. The second figure delineates the class-wise accuracy curves for CNN-M. The experimental statistics are based on the SiftFlow dataset \cite{liu2009nonparametric}.}
  \label{Fig:siftFlow-sampling}
\end{figure}
Knowing that the number of training patches are prohibitively huge (thousands of millions), we only use a fraction of them for the network training. Specifically, at the beginning of each epoch, training patches are randomly sampled. By doing this, we are able to decrease the training time dramatically. Meanwhile, the sampling strategy does not harm the performance, as the image patches are highly redundant (patches that are related to neighborhood pixels are usually the same) and the randomness injected into the data sampling during each epoch enables the network to ``see'' the whole data throughout the whole training process.  Here in this paper, we introduce four sampling methods based on the class frequency distribution in natural scene images:

\begin{itemize}
  \item \textbf{Global sampling (GS)}: It samples patches randomly from the collection of all the training patches. Then the class frequency distribution in the sampled patches should be very close to that in the original images.
  \item \textbf{Class sampling (CS)}: It samples patches in a manner that classes appear equally in the resulting sampled patches. Note that some patches may appear multiple times if they are from extremely rare classes.
  \item \textbf{Hybrid sampling (HS)}: It is a compromise between global sampling and class sampling. In detail, it firstly samples patches globally, and then augments the rare-class patches until their occurrence frequencies reach the desired threshold $\eta$ in the sampled patches.
  \item \textbf{Truncated class sampling (TCS)}: It adopts the same sampling procedure as class sampling, but removes all the frequent-class patches.
\end{itemize}

The threshold $\eta$ is used to determine whether a class is frequent or rare: the class is considered to be frequent as long as its occurrence frequency in the training data is above $\eta$, otherwise, it belongs to a rare class.
Obviously, the above sampling methods are expected to yield significantly different sampled patches in the form of class frequency distribution.
When these image patches are presented in the network training phase, the CNNs are enforced to learn disparate visual patterns, thus they behave differently.
We next discuss how the sampled image patches influence the characteristics of CNNs.

\subsection{CNN-Ensemble}
CNNs, even with the identical architecture, can be functionally different if they are trained from image patches with disparate data distribution. Concretely, to fit the image patches whose statistics are not the same, CNNs adaptively learn different representations and classifiers. In consequence, it leads to the big performance discrepancy on labeling the same images. Let's denote \textbf{CNN-M} the CNN trained with image patches sampled through method M. As shown in Figure \ref{Fig:siftFlow-sampling}, CNN-M exhibits significantly different characteristics towards predicting the semantic classes of pixels: \textbf{CNN-GS} prioritizes the frequent classes, therefore it optimizes the overall pixel accuracy; \textbf{CNN-CS} implicitly normalizes the weights of classes by performing downsampling and oversampling operations to frequent and rare classes respectively, thus it maximizes the average class accuracy; In comparison with CNN-GS, \textbf{CNN-HS} gives slightly higher weights to rare classes,  hence it compromises the two above criterions; \textbf{CNN-TCS} works exceptionally well on differentiating rare classes.

To produce a satisfactory labeling prediction map, the algorithm is required to perform extraordinarily well to correctly predict frequent classes, and in the mean time should work well towards recognizing rare classes precisely. The former criterion guarantees that the scene semantics of the images are well defined, and the latter one enforces that objects (rare classes) are not missing in the scene.  However as manifested in Figure \ref{Fig:siftFlow-sampling}, none of the CNN-M satisfies both criterions. Therefore, we propose the \textbf{CNN-Ensemble}, which combines the predictions from the CNN-M components. Mathematically, the local belief $P_I(X_i,Y_j)$ is derived with the following equation:
\begin{equation}
P_I(X_i,Y_j)= \frac{1}{M}\sum_{m=1}^{M}P_m(X_i, Y_j)
\end{equation}
where $M$ is the number of CNN-M components and $P_m(X_i, Y_j)$ denotes the class likelihood prediction from the m-th CNN-M.
As mentioned, CNN-M components are designed to focus on distinguishing some specific classes. For example, CNN-GS discriminates frequent classes excellently, and CNN-TCS captures the subtleties of infrequent classes. The proposed CNN-Ensemble fuses the complementary predictions from different sources. Our later experiments in Section \ref{Section:Experiment} will demonstrate that CNN-Ensemble is able to produce much more reliable labeling results quantitatively, and it performs excellently on both overall pixel accuracy and average per-class accuracy.

%

\section{Global Beliefs}
\label{Section:global_belief}

In a global view, the semantic class of each pixel $X_i$ in image $X$ is determined by the global scene semantics of $X$. In other words, its global class likelihood $P_G(X_i, Y_j)$ should match the expected scene layout of $X$. Even though the pixel $X_i$ is represented identically, its class belief could vary significantly depending on how the pixel is evaluated. Take the simple example as an illustration in Figure \ref{Fig:intro}, the pixels in the lower part of the image could be `sand', `road' or even `rock' in a local view. We refer to this problem as \textit{local ambiguity}. In contrast, with the awareness of global `coast' scene, it is obvious that `sand' class is preferred in a global view. In this perspective, the global scene prior is a good remedy to alleviate the local ambiguity.

\subsection{Non-parametric Global Belief Transfer}
As elaborated, the parametric CNNs (CNN-Ensemble) are able to produce good labeling results for the pixels with good local contextual support, they still suffer from the notorious \textit{local ambiguity} problem.
Previously, researchers usually addressed this issue by generating contextual aware local features. For example, Farabet et al. \cite{farabet2013learning} fed the network with multi-scale image patches to yield richer contextual aware local features, and likewise Pinheiro et al. \cite{pinheiro2014recurrent} took the network input as larger image patches. In this paper, the local disambiguation is achieved by enforcing a global scene constraint to local classification.
More specifically, a pixel is considered under global context: the class likelihood of pixels should satisfy the scene layout and semantics of the image.

\textcolor{black}{First, we generate the global-level representation for the considered image $X \in \mathbb{R}^{h \times w \times 3}$, where $h,w$ is the height and width of the image $X$. The corresponding CNN feature tensor
$F\in\mathbb{R}^{h\times w\times M}$ can be obtained by passing densely sampled image patches in $X$ to the truncated CNN ($M = 64$ in our implementation).
Next, we introduce the average pooling operator $pool$ \cite{krizhevsky2012imagenet} to aggregate the pixel features, thus giving rise to the global feature $H$.
In detail, suppose an image is decomposed to regions  $\mathcal{R}=\{R^{(1)},R^{(2)},\ldots,R^{(J)}\}$ \footnote{$J$ is the number of regions. In our experiments, the image is divided into rectangular regions in a 2-layer spatial pyramid fashion \cite{lazebnik2006beyond}.}, } the region feature is generated by applying $pool$ operator to the constituent pixel features: $H(R^{(i)}) = pool(F^{i}), \forall i \in R^{(i)} $. The global image representation is defined as the concatenation of region features $H = [H(R^{(1)}),H(R^{(2)}),\ldots,H(R^{(J)})]$. As expected, this global image feature $H$ not only conveys discriminative scene semantics but also encodes scene layout information.

Then, based on $H$, the global nearest exemplars $\mathcal{S}(X)$  are retrieved.  Each image in $\mathcal{S}(X)$  is expected to have the similar scene semantics and layout with image $X$. After that, the global class likelihood of pixels (global belief) are transferred from the statistics of pixel features in $\mathcal{S}(X)$. \textcolor{black}{Concretely, among all the pixels in $\mathcal{S}(X)$, the semantic class of pixel $X_i$ should match those pixels whose local representations are also close to $X_i$. Therefore, $K$ pixels (from images within $\mathcal{S}(X)$) are firstly retrieved that are similar to $X_i$ in the local representation space. Then the global belief is generated through a weighted voting based on the $K$ retrieved pixels.} Mathematically, it is derived in the following equation:
\begin{equation}
P_G^{\mathcal{S}(X)}(X_i,Y_j) = \frac{\sum_{k=1}^{K}\phi(X_i,X_k)\delta(Y(X_k)=Y_j)}{\sum_k\phi(X_i,X_k)}
\label{Equation:knn}
\end{equation}
where $X_k$ is the $k$-th nearest neighbor pixel of $X_i$ among all the pixels in $\mathcal{S}(X)$; $Y(X_k)$ is the ground truth label for pixel $X_k$; $\delta(Y(X_k),Y_j)$ is an indicator function;  $\phi(X_i,X_k)$ measures the similarity between $X_i$ and $X_k$, which is defined over spatial and feature space:
\begin{equation}
\begin{aligned}
\phi(X_i,X_j) = exp(-\alpha||x_i-x_j||)exp(-\gamma||z_i-z_j||)\\
\end{aligned}
\label{Equation:feature_similarity}
\end{equation}
where $x_i=F(X_i)$ denotes the CNN pixel feature for $X_i$, $z_i$ is the normalized coordinate along the image height axis and $\alpha, \gamma$ controls the belief exponential falloff.

Meanwhile, as small-size object classes (e.g. `bird' in the sky, `boat' in the sea, etc) make negligible contribution to the global scene semantics, they may not appear in its nearest global exemplar images $\mathcal{S}(X)$. Thus, the global belief will be highly skewed to frequent classes, which potentially harms the final class likelihood of rare classes according to the integration rule in Equation \ref{Equation:main}. To address this issue, we introduce an auxiliary \textcolor{black}{pixel transfer set} $\mathcal{A}(X)$. In detail, we first find the rare-class \textcolor{black} {pixels} whose quantities are below $K$ (same value as in Equation \ref{Equation:knn}) in $\mathcal{S}(X)$, and then augment the corresponding rare-class \textcolor{black}{pixels} until their quantities reach $K$. \textcolor{black}{More specifically, $\mathcal{A}(X)$ is a set of rare-class pixels, and it is derived by randomly sampling the desired quantity of pixels from images outside $\mathcal{S}(X)$. } Thus, the final transfer set is the combination of $\mathcal{S}(X)$ and $\mathcal{A}(X)$, and the quantity of every rare-class pixel is at least $K$. Thus, the global belief is expected to preserve the salient objects in the scene.
Our non-parametric global belief estimation is reminiscent of popular label transfer works\cite{eigen2012nonparametric,liu2009nonparametric,singh2013nonparametric,tighe2010superparsing,tung2014collageparsing}, two differences need to be highlighted:

\begin{itemize}
\item Instead of adopting hand-engineered low-level local and global features, we use more discriminative and compact features learned from CNN for label transfer.

\item Our non-parametric model works as global scene constraints for local pixel features. Generally, small size retrieval images are sufficient to define the scene semantic and layout. However, previous works have to seek large retrieval set to cover all the possible semantic classes.

\end{itemize}

\subsection{CNN-Metric}
\begin{figure}[t]
\begin{center}
\includegraphics[width=0.35\textwidth]{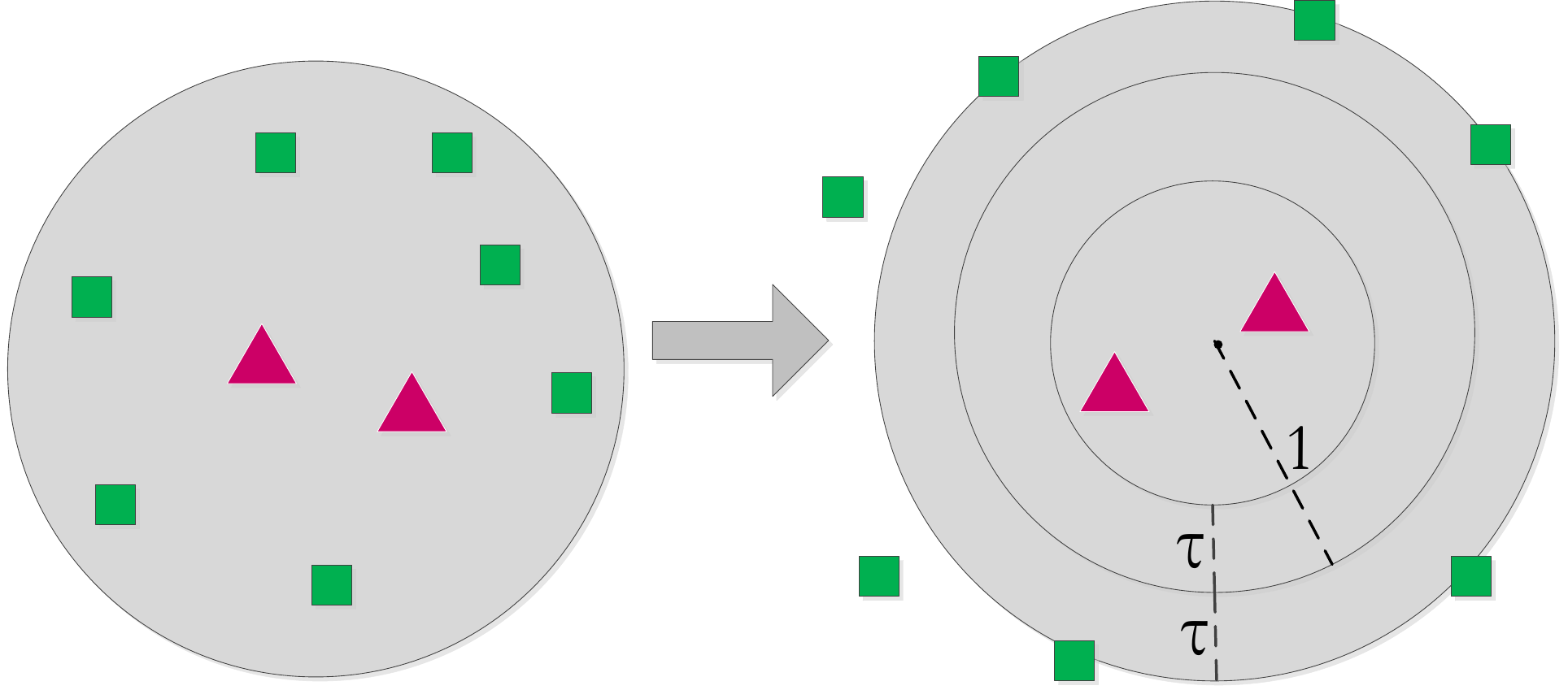}
\end{center}
\caption{Graphical illustration of the effect of our large margin based metrics. Due to the highly imbalanced data distribution, the nearest neighbors of the testing feature (triangle) are dominated by imposter classes (rectangle). After the metric transformation, the imposters stay far away from it. Thus, their contribution is significantly attenuated in the global belief estimation.}
\label{Fig:wknn}
\end{figure}

As shown in Equation \ref{Equation:knn}, the estimation of global belief $P_G^{\mathcal{S}(X)}(X_i,Y_j)$ is highly dependent on the distance metric between two pixel features. However, our features are learned by optimizing pixel/patch classification accuracy, while do not take distance metric into consideration.
Therefore we propose to learn a large-margin based metric to mitigate the inaccurate class likelihood estimation for rare classes (Figure \ref{Fig:wknn}). In detail, the Mahalanobis metric $M = W^TW$ is learned by minimizing the loss function, which  is formally written as:
\begin{equation}
\begin{aligned}
 &L = \frac{\lambda}{2}||W||^2 + \frac{1}{2N}\sum_{i,j}g(x_i,x_j) \\
 &g(x_i,x_j) = max(0,1-\ell_{i,j}(\tau-||Wx_i-Wx_j||^2))
 \end{aligned}
 \label{Equation:lmcnn}
\end{equation}
where $\ell_{i,j}$ indicates whether two features have the same semantic label or not, and $\ell_{i,j}=1$ if $X_i$ and $X_j$ are from the same class, or $\ell_{i,j} = -1$ otherwise; $\tau(>1)$ is the margin and $\lambda$ controls the effect of regularization; $x_i=F(X_i)$ is the feature representation for $X_i$ and $N$ is the number of features. The objective function would enforce the pixel features from the same semantic class to be close and stay within the ball with radius $1-\tau$, and enforce data from different classes to be far away from each other by at least $1+\tau$. The graphical illustration of the metrics is depicted in Figure \ref{Fig:wknn}.

Instead of simply learning a metric based on the extracted CNN features, we further replace the softmax layer with our metric learning layer, so that the feature extraction parameters can also be adapted.  We replace the softmax layer of previous CNN (CNN-softmax) with a fully connected layer parameterized by $W$  (or more layers to learn non-linear metrics \cite{hu2014discriminative}) and fix the biases to be zero, which serves as a Mahalanobis metric ($M = W^TW$). We call the new network CNN-metric. These two networks do not share any parameters except that the feature extraction parameters of CNN-metric are initiated from the corresponding layers of CNN-softmax. The errors are back propagated through the chain rule, and $\frac{\partial L}{\partial W} \ \frac{\partial L}{\partial x_i}$ for the last layer are given in Equation \ref{Equation:bp}.

\begin{equation}
\begin{aligned}
&\frac{\partial L}{\partial W} = \lambda W + \frac{1}{N}\sum_{i,j}\zeta_{ij}\\
&\zeta_{ij} = g^{'}(c)\ell(i,j)(Wx_i-Wx_j)(x_i-x_j)^T\\
&\frac{\partial L}{\partial x_i} = \frac{1}{N}\sum_{i,j}g^{'}(c)(W^T\ell(i,j)(Wx_i-Wx_j))\\
&c = 1 - \ell(i,j)(\tau-||Wx_i-Wx_j||^2)\\
&x_i = F(X_i)\\
&g^{'}(c) = \left\{ \begin{array}{ll}
   0, & c <= 0  \\
   1, & c > 0 \end{array} \right.
\end{aligned}
\label{Equation:bp}
\end{equation}

We adopt the Stochastic Gradient Descent (SGD) to optimize the CNN-Metric. Considering that the quantity of patches is prohibitively huge, we also sample a fraction of patches during each epoch in the \textbf{Class Sampling} manner (Section \ref{Section:local_belief}). Other sampling methods are not appropriate in this scenario, as the imbalanced data distribution will result in the scarcity of training examples for some class pairs, which is expected to skew the learned metric mapping. Furthermore, due to the infeasibility of feeding the sampled patches to the network in a single propagation, the training data are divided to several batches, among which the proposed metric constraint is enforced to any class pairs.


\section{Experiments}
\label{Section:Experiment}
\subsection{Evaluation Benckmarks}
We evaluate our approach on two popular and challenging scene parsing benchmarks:
\begin{itemize}
  \item SiftFlow \cite{liu2009nonparametric}: It has 2688 images generally captured from 8 typical natural scenes. Every image has 256 $\times$ 256 pixels, which belongs to one of 33 semantic classes. We use the training/testing (2488/200 images) split provided by \cite{liu2009nonparametric} to conduct our experiments.
  \item Barcelona \cite{tighe2010superparsing}: It consists of 14871 training and 279 testing  images. The size of the images varies across different instances, and each pixel is labelled as one of the 170 semantic classes. Note that the class frequency distribution is more imbalanced than that in the SiftFlow dataset. Meanwhile, the scene categories of training images range from indoor to outdoors, whereas the testing images are only captured from the barcelona street.
      These issues pose Barcelona as an extremely challenging dataset.
\end{itemize}
To quantitatively evaluate our methods, we report two types of scores: the percentage of all correctly classified pixels - Global Pixel Accuracy (\textbf{GPA}), and Average per-Class Accuracy (\textbf{ACA}).
\subsection{Local Labeling Results}

\begin{figure}[t]
\begin{center}
  \includegraphics[width=0.24\linewidth]{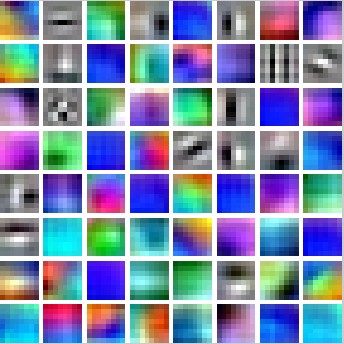}
  \includegraphics[width=0.24\linewidth]{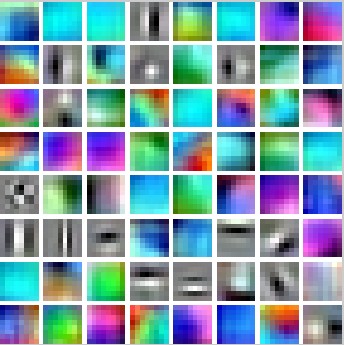}
  \includegraphics[width=0.24\linewidth]{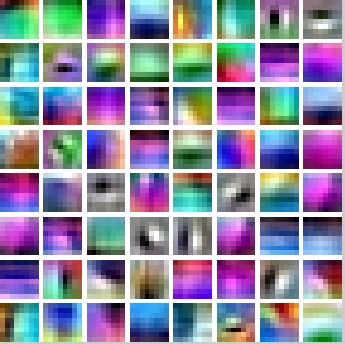}
  \includegraphics[width=0.24\linewidth]{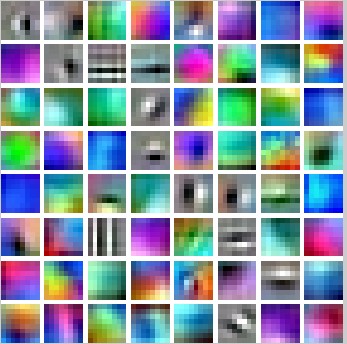}
\end{center}
\caption{Visualization of the learned convolution filters for the first layer of CNN-CS, CNN-TCS, CNN-GS and CNN-HS respectively. They are trained on the Barcelona datasets. Note the visual difference among the learned filters of different CNN-M. The figure is best viewed in color.}
\label{Fig:filters_visualization}
\end{figure}

We first present the implementation details of training the parametric CNNs. We use Stochastic Gradient Descent (SGD) with momentum to train the CNN-M; During each training epoch, around $5\times10^{5}$ pixels are randomly sampled from training pixel pools; The learning rate is initialized to be 0.01, and it is decreased by 10 times after 20 epoches; The momentum is fixed to 0.9, and the batch size is set as 100. We train our CNN-M based on MatConvnet \cite{arXiv:1412.4564} toolbox. \footnote{The code is publicly available under the homepage of authors.} The reported results are based on the model trained in 35 epoches. Each image is preprocessed by first subtracting the mean and then performing contrast normalization by dividing its variance.
\textcolor{black}{Besides, the threshold $\eta$ that discriminates rare classes are empirically set to $5\%$ and $1\%$ for the SiftFlow and Barcelona datasets respectively.}
The learned convolutional filters for the first layer of CNN-M on the Barcelona dataset are shown in Figure \ref{Fig:filters_visualization}.


\begin{table}[t]
\begin{center}
\begin{tabular}{|l|cc|cc|}
\hline
& \multicolumn{2}{c|}{SiftFlow}  & \multicolumn{2}{c|}{Barcelona} \\
& GPA & ACA & GPA & ACA \\
\hline
Multiscale ConvNet \cite{farabet2013learning} & 67.9\% & 45.9\% & 37.8\% & 12.1\% \\
Recurrent CNN (67$\times$67) \cite{pinheiro2014recurrent}& 65.5\% & 20.8\% & N/A & N/A\\
\hline
\textcolor{black}{CNN-asymmetric} \cite{mostajabi2015feedforward}& 42.4\% & 38.4\% & 20.0\% & 13.3\% \\
\hline
\hline
CNN-GS & 75.4\% & 30.2\% & 68.5\% & 11.4\%\\
CNN-CS & 70.9\% & 42.6\% & 24.7\% & 18.4\% \\
CNN-TCS & 7.91\% & 38.7\% & 6.33\% & 16.8\% \\
CNN-HS & 74.7\% & 39.4\% & 61.0\% & 16.7\% \\
\hline
\hline
CNN-Ensemble & 75.3\% & \textbf{44.8\%} & 61.3\% & 19.5\% \\
\hline
\hline
Ensemble CNN-GS& \textbf{77.1}\% & 32.0\% & \textbf{69.7}\% & 11.5\% \\
Ensemble CNN-CS& 72.8\% & 43.7\% & 26.6\% & \textbf{20.0}\% \\
Ensemble CNN-TCS& 8.33\% & 40.0\% & 7.53\% & 18.3\% \\
Ensemble CNN-HS& 76.4\% & 40.2\% & 63.2\% & 17.5\% \\
\hline
\end{tabular}
\end{center}
\caption{Quantitative Performance of different CNNs. Details of each method are elaborated in the text.}
\label{Table:cnn-result-comp}
\end{table}

Next, we evaluate the performance of CNN-M. Specifically, CNNs output their local belief maps, based on which maximum marginal inference is performed to output the label prediction maps. Table \ref{Table:cnn-result-comp} lists the quantitative results. As shown, our CNN-M (GS, HS, CS) achieves much better results than Multiscale ConvNet and Recurrent CNN. In terms of the individual performance of CNN-M,
\textbf{CNN-GS} achieves the best accuracy on global pixel accuracy (GPA), whereas its performance on average class accuracy (ACA) is not satisfactory. In contrast, \textbf{CNN-CS} claims the best ACA among all the single CNNs, which indicates that it predicts the semantic classes of pixels in a more equal manner. A  favorable performance compromise is achieved by \textbf{CNN-HS}, which works considerably well on both GPA and ACA. \textbf{CNN-TCS} performs extremely poor on GPA as it ignores the frequent classes, it however achieves very competitive ACA.  Importantly, CNN-TCS captures image statistics that are significantly disparate from other networks (CNN-GS, HS, CS): it performs the best to correctly recognize the rare-class pixels. Quantitatively, the CNN-Ensemble that excludes CNN-TCS only achieves 40.3\% and 17.8\% in terms of ACA on the SiftFlow and Barcelona dataset respectively, and their performance are boosted to 44.8\% and 19.5\% after including CNN-TCS. By fusing the complementary decisions from different CNN-M, \textbf{CNN-Ensemble} achieves the best performance tradeoff.

Furthermore, we train \textbf{Ensemble CNN-M} to compare their performance behaviour with CNN-Ensemble. In detail, the CNN-M in Ensemble CNN-M is trained with the image patches sampled from the identical method M. As manifested in Table \ref{Table:cnn-result-comp}, the Ensemble CNN-M significantly improves the local labeling performance over CNN-M. However, they optimize only one evaluation criterion (either GPA or ACA), and this phenomenon is more obvious in the severely imbalanced Barcelona dataset.  In contrast, the CNN-Ensemble performs competitively excellently on both GPA and ACA. It's also important to note that CNN-HS behaves similarly with CNN-Ensemble and it produces very promising results on both GPA and ACA. However, the performance of Ensemble CNN-HS is significantly inferior to CNN-Ensemble in terms of ACA.

\textcolor{black}{
In the end, we discuss the issue of imbalanced class frequency distribution in scene parsing.
In order to boost the recognition rates for infrequent classes, we train another \textbf{CNN-asymmetric} as in \cite{mostajabi2015feedforward}, whose log-loss is modulated by the inverse frequency of each class, thus the rare classes are effectively given more attention. In this case, the training data can be generated as in \cite{mostajabi2015feedforward} by collecting all image patches (or via global sampling strategy).
The corresponding result is reported in Table \ref{Table:cnn-result-comp}, which shows that CNN-asymmetric fails to achieve the desirable results as in object segmentation benchmarks \cite{mostajabi2015feedforward}.
This phenomenon can be explained from the following perspective. The class frequency distribution in the object segmentation task is not as imbalanced as in the scene parsing task \footnote{Statistically, the frequency ratio between the most frequent and rare classes on the PASCAL VOC 2011 \cite{mostajabi2015feedforward} and SiftFlow datasets (see Figure \ref{Fig:siftFlow-sampling}) are approximately $240$ and $3.5 \times 10^4$ respectively. }.
If the inverse frequency is used to scale the log-loss in this scenario, the scaled losses w.r.t the frequent classes are negligible.
Consequently, the CNN-asymmetric performs poorly on GPA on the scene parsing benchmarks.
In contrast, the sampling based CNNs (e.g. CNN-CS, CNN-HS, CNN-GS) achieve much better performance tradeoff, which positively elucidates the effectiveness of the proposed sampling strategy to address the class imbalance issue. More importantly, by feeding the networks with different sampled patches during the training phase, we are able to train a number of complementary CNNs, and combine them to produce much stronger local prediction model (CNN-Ensemble).}




\subsection{Evaluation of Global Features}
\begin{table}
\begin{center}
\begin{tabular}{|l|cccc|}
\hline
 & Dim & K=1 & K=5 & K=10 \\
 \hline
 GIST \cite{oliva2006building} & 512 & 74.0\% & 70.7\% & 68.3\% \\
 SIFT-SPM \cite{lazebnik2006beyond} & 2100 & 76.5\% & 71.3\% & 69.1\% \\
 \hline
 \hline
 Global Feature (CNN-GS) & \textbf{320} & \textbf{91.5}\% & \textbf{88.3}\% & 86.5\% \\
 Global Feature (CNN-HS) & \textbf{320} & 88.5\% & 85.8\% & 85.2\% \\
 Global Feature (CNN-CS) & \textbf{320} & 90.5\% & 84.7\% & 84.1\% \\
 Global Feature (CNN-TCS) & \textbf{320} & 79.5\% & 73.9\% & 72.8\% \\
 Global Feature (CNN-Ensemble) & 1280 & \textbf{91.5}\% & 87.4\% & \textbf{86.7}\% \\
 \hline
 \hline
 GT & 165D & 94.0\% & 91.0\% & 89.5\% \\
 \hline
 \end{tabular}
\end{center}
 \caption{Average genuine matching percentage among their K-nearest neighbors for different global features. GT is the semantic feature pooled from ground truth label maps. }
 \label{Table:globalFeature}
\end{table}
In this section, we demonstrate that the pooling operation of pixel features is capable of generating semantically consistent global features. To achieve this goal, we calculate the KNN matching score $p$ - the average genuine matching percentage among their $K$ nearest neighbors. It is Mathematically derived in the following equation: $p = \frac{\sum_i^N\sum_k^K\delta(I_i, NN(i,k))}{NK}$, where $N$ is the number of test images, $NN(i,k)$ stands for the $k$-th nearest neighbor for image $I_i$, and $\delta(i,j)$ outputs value 1 if $i$ and $j$ are a genuine match, or 0 otherwise. A genuine matching image pair means that they belong to the identical semantic class. We test the global features on the SiftFlow dataset, as it provides the global scene label for each image.

Four global features are compared in our experiment: GIST \cite{oliva2006building} is a global summary of scene images that captures scene structure and layout; SIFT-SPM (GT) \cite{lazebnik2006beyond} is pooled from low-level local SIFT \cite{lowe2004distinctive} (ground truth label map \cite{liu2009nonparametric})  in a 3(2)-layer spatial pyramid. They are commonly used in scene classification and non-parametric label transfer framework. GT is the ideal global semantic feature. As mentioned in Section \ref{Section:global_belief}, our global feature is pooled from the output of truncated CNN-M in a 2-layer spatial pyramid fashion. Euclidean distance is used to retrieve nearest neighbors for non-histogram features (GIST, Ours), and histogram intersection similarity measurement is applied for the rest histogram features (SIFT-SPM and GT).

The quantitative matching scores for different global features are listed in Table \ref{Table:globalFeature}. As demonstrated, our global feature is more likely to group semantically relevant images together. Meanwhile, among all of the global features(CNN-M), CNN-GS performs the best and CNN-TCS works the worst, as the scene semantics of outdoor images are mostly determined by the appearance of frequent classes. It's worth noting that the concatenation of global features from different CNN-M fails to outperform CNN-GS. Hence, only global feature (CNN-GS) is used subsequently to retrieve the nearest exemplars. It's also interesting to observe that the retrieval performance based on GT features are imperfect, which implies that different scenes can have very similar building blocks. For example, `inside city' and `street' scenes are dominated by sky and building pixels. We believe that the quality of nearest neighbor retrieval directly determines the correctness of global belief. Therefore, our global feature is expected to benefit other label transfer works as well. Some qualitative examples are shown in Figure \ref{Fig:qualitative_result}, in which the retrieved nearest exemplar images have very similar scene layout.

\subsection{Global Labeling Results}

We first present the implementation details for our non-parametric model. The non-overlapping patches are adopted as label transfer units, within which labels are assumed to be identical. Specifically, as our CNN-M has three pooling layers, the feature extractor (truncated CNN-M) can be regarded as sliding the images with a stride of 8.
In other words, the dimension of the output feature tensor $F$ is $\frac{1}{8}$ of original image size: one feature in $F$ corresponds to a 8$\times$8 image patch. In our experiments, we only estimate the class likelihood for each feature in $F$, implicitly assuming that class labels within the $8\times8$ regions are the same.\footnote{This is not the optimal setting, as it oversmoothes the label prediction maps. However, it is fast and easy to implement, and we don't expect that the global belief to preserve the boundary information as it simply reflects the global scene prior. Hence, it's a reasonable compromise.}

The global belief is calculated by Equation \ref{Equation:knn}, in which $|\mathcal{S}(X)|$ (size of nearest exemplar images) and $K$ (size of nearest pixel/patch neighbors) are empirically set to be 5 and 200 respectively. As manifested in Table \ref{Table:globalFeature}, $5$ images are sufficient to correctly define the scene semantics. However, as many images are not fully annotated in the Barcelona dataset, a larger retrieval image set is used ($|\mathcal{S}(X)|=100$). $\alpha$ and $\gamma$ in Equation \ref{Equation:feature_similarity} are empirically set to $15$ and $5$ respectively.
To fine-tune the CNN-metric, $1 \times 10^{4}$ patches are sampled for each class in each epoch (which results in $3.3\times 10^{5}$ and $1.7 \times 10^{6}$ training patches on the SiftFlow and Barcelona dataset respectively), and $2000$ patches are used to be a training batch. $\lambda$ and $\tau$ in Equation \ref{Equation:lmcnn} makes marginal difference to the performance, and they are fixed to $0.01$ and 3 respectively. The learning rate is initialized to $10^{-3}$ and decays exponentially with the rate of $0.9$. The reported results are obtained under the models learned in 20 epoches.

\begin{table}[t]
\begin{center}
\begin{tabular}{|l|c|c|}
\hline
 & Global (GPA) & Class (ACA) \\
\hline
SuperParsing \cite{tighe2010superparsing} & 76.9\% & 29.4\% \\
Liu et al. \cite{liu2009nonparametric} & 74.8\% & N/A \\
Gould et al. \cite{gould2014superpixel} & 78.4\% & 25.7\% \\
Singh et al. \cite{singh2013nonparametric} & 79.2\% & 33.8\% \\
Tighe et al. \cite{tighe2013finding} & 78.6\% & 39.2\% \\
Yang et al. \cite{yangcontext} & 79.8\% & 48.7\% \\
\textcolor{black}{Tung et al.} \cite{tung2015scene} & 79.9\% & 49.3\% \\
\textcolor{black}{George et al.} \cite{george2015image} & \textbf{81.7}\% & \textbf{50.1}\% \\
\hline
\hline
Raw Multiscale ConvNet \cite{farabet2013learning} & 67.9\% & 45.9\% \\
Raw Multiscale ConvNet \cite{farabet2013learning} + Cover & 72.3\% & 50.8\% \\
Raw Multiscale ConvNet \cite{farabet2013learning} + Cover & 78.5\% & 29.4\%\\
Plain CNN (133$\times$133) \cite{pinheiro2014recurrent} & 76.5\% & 30.0\% \\
Recurrent CNN (133$\times$133) \cite{pinheiro2014recurrent} & 77.7\% & 29.8\% \\
Gatta et al. \cite{gattaunrolling} & 78.7\% & 32.1\% \\
\textcolor{black}{Long et al.} \cite{long2015fully} (ImageNet Pretrain) & 85.2\% & 51.7\% \\
\hline
\hline
Local Labeling (CNN-Ensemble) & 75.3\% & 44.8\% \\
Global Labeling & 78.7\% & 36.2\% \\
Global Labeling (Metric) & 78.8\% & 39.6\% \\
Integration model & 81.0\% & 44.6\% \\
Integration model (Metric) & \textbf{81.2}\% & \textbf{45.5}\% \\
\hline
\end{tabular}
\end{center}
\caption{Quantitative performance of different methods on the SiftFlow datasets.}
\label{Table:integration-result-siftFlow}
\end{table}

\begin{table}[t]
\begin{center}
\begin{tabular}{|l|c|c|}
\hline
 & Global (GPA) & Class (ACA) \\
 \hline
 SuperParsing \cite{tighe2010superparsing} & 66.9\% & 7.6\% \\
 \hline
 \hline
 Raw Multiscale ConvNet \cite{farabet2013learning} & 37.8\% & 12.1\% \\
 Raw Multiscale ConvNet \cite{farabet2013learning} + Cover & 46.4\% & \textbf{12.5}\% \\
 Raw Multiscale ConvNet \cite{farabet2013learning} + Cover & \textbf{67.8}\% & 9.5\% \\
 \hline
 \hline
 Local Labeling (CNN-Ensemble) & 61.3\% & 19.5\% \\
 Global Labeling & 68.1\% & 13.1\% \\
 Global Labeling (Metric) & 68.7\% & 14.1\% \\
 Integration Model & 69.7\% & 16.8\% \\
 Integration Model (Metric)& \textbf{70.3}\% & \textbf{17.2}\% \\
 \hline
\end{tabular}
\end{center}
\caption{Quantitative performance of different methods on the Barcelona dataset.}
\label{Table:integration-result-barcelona}
\end{table}

Similarly, the global labeling results are obtained through performing maximum marginal inference over global beliefs. The quantitative results on the SiftFlow and Barcelona dataset are listed in Table \ref{Table:integration-result-siftFlow} and \ref{Table:integration-result-barcelona} respectively.
From which, we clearly observe that our simple non-parametric model achieves very promising results that are comparable or even better than most complicated label transfer counterparts. For example, we reach 78.8\% (39.6\%) in terms of GPA (ACA) on the SiftFlow dataset, whereas SuperParsing \cite{tighe2010superparsing} and graph transfer \cite{gould2014superpixel} only achieves 76.9\% (29.4\%) and 78.4\% (25.7\%) respectively.  Moreover, our global labeling alone already achieves state-of-the-art results on the challenging Barcelona dataset. We attribute the performance superiority to the highly discriminative CNN features which we work with in the non-parametric transfer model.

Furthermore, as evidenced by Table \ref{Table:integration-result-siftFlow} and \ref{Table:integration-result-barcelona}, the learned metric is capable of improving the quality of global beliefs for rare classes. As illustrated in Figure \ref{Fig:wknn}, the learned metric is expected to shrink the distance of pixel features between identical classes and enlarge the distance between disparate classes. Consequently, it benefits the estimation of global beliefs for infrequent classes. In our experiments, we do observe the desirable average class accuracy (ACA) boost by incorporating metric tuning on both datasets.

\subsection{Integration Labeling Results}

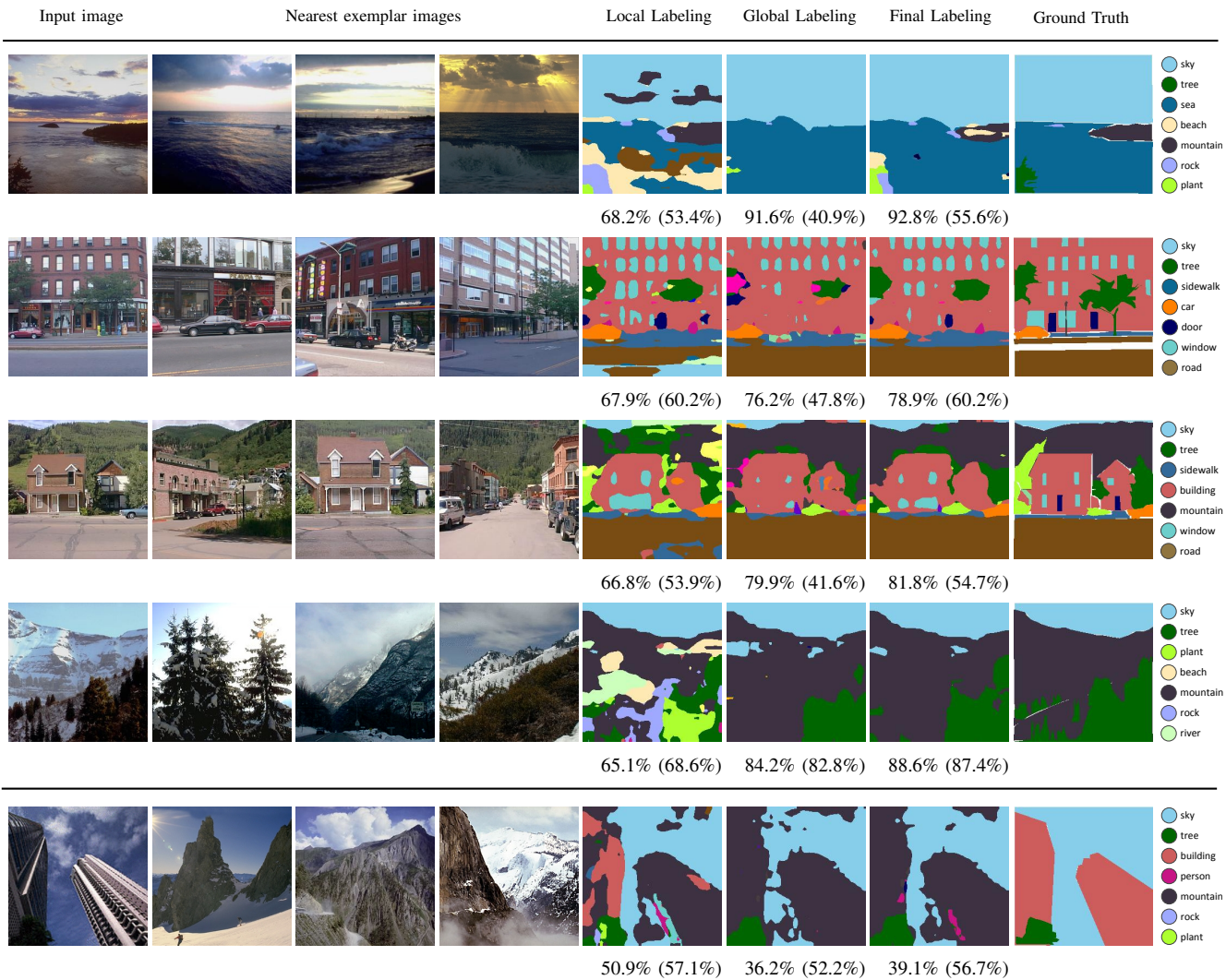
\begin{figure*}
\newcommand{\InsertImageS}[1]{
\begin{subfigure}[t]{0.11\textwidth}
\centering
\includegraphics[height=\textwidth]{#1}
\end{subfigure}
\hspace{-0.02\textwidth}
}
\newcommand{\InsertImageD}[3]{
\begin{subfigure}[t]{0.11\textwidth}
\captionsetup{
justification=raggedleft,singlelinecheck=false, position=top
}
\centering
\includegraphics[height=\textwidth]{#1}
\caption*{\footnotesize{{#2 (#3)}}}
\end{subfigure}
\hspace{-0.02\textwidth}
}

\begin{minipage}{0.115\textwidth}
\centering
\scriptsize{Input image}
\end{minipage}
\begin{minipage}{0.335\textwidth}
\centering
\scriptsize{Nearest exemplar images}
\end{minipage}
\begin{minipage}{0.105\textwidth}
\centering
\scriptsize{Local Labeling}
\end{minipage}
\begin{minipage}{0.105\textwidth}
\centering
\scriptsize{Global Labeling}
\end{minipage}
\begin{minipage}{0.105\textwidth}
\centering
\scriptsize{Final Labeling}
\end{minipage}
\begin{minipage}{0.105\textwidth}
\centering
\scriptsize{Ground Truth}
\end{minipage}

\vspace{-0.5em}

\begin{picture}(1,1)
\setlength{\unitlength}{\textwidth}
\thicklines
\put(-0.005,0){\line(1,0){0.965}}
\end{picture}

\begin{center}
\InsertImageS{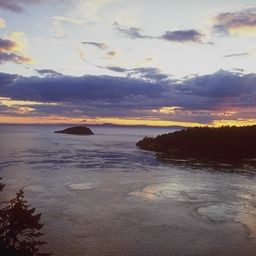}
\InsertImageS{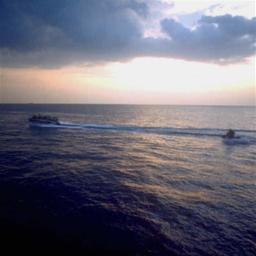}
\InsertImageS{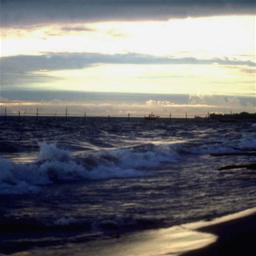}
\InsertImageS{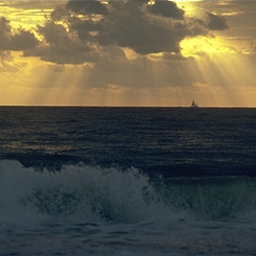}
\InsertImageD{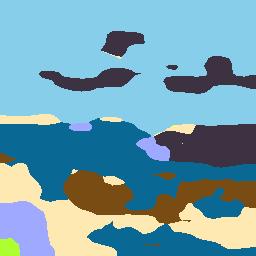}{68.2\%}{53.4\%}
\InsertImageD{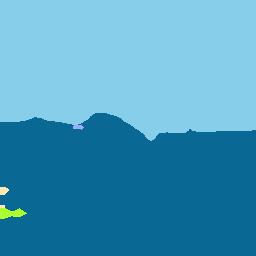}{91.6\%}{40.9\%}
\InsertImageD{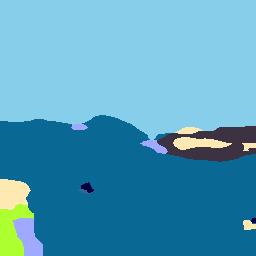}{92.8\%}{55.6\%}
\InsertImageS{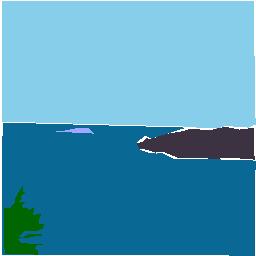}
\hspace{-1.75em}
\InsertImageS{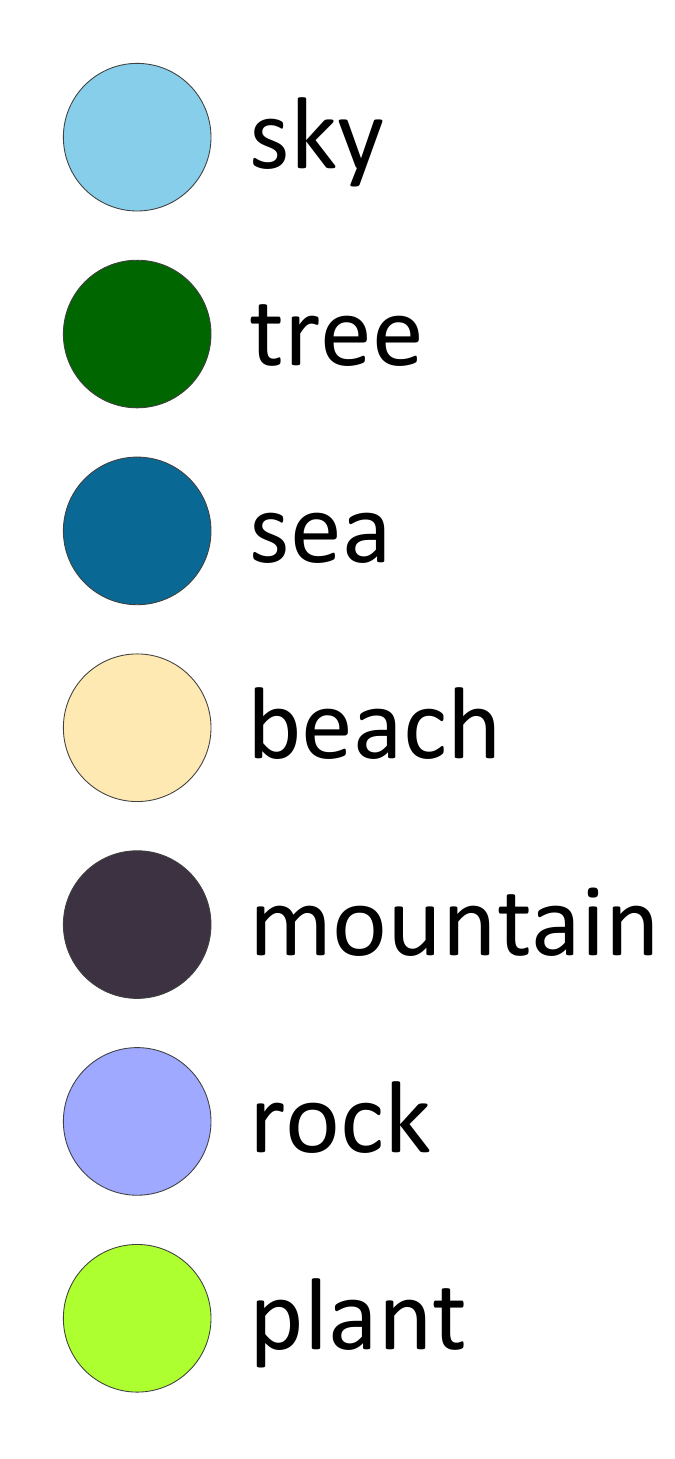}\\
\vspace{-0.25em}



\InsertImageS{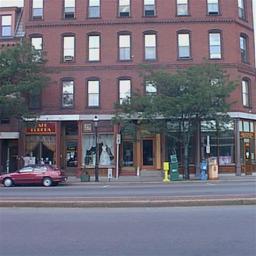}
\InsertImageS{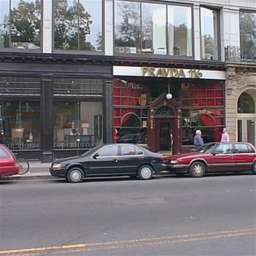}
\InsertImageS{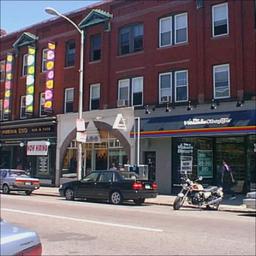}
\InsertImageS{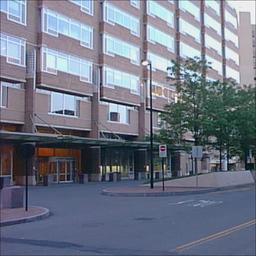}
\InsertImageD{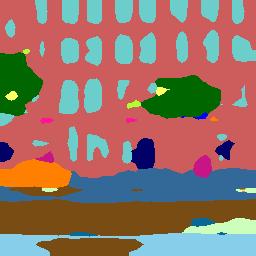}{67.9\%}{60.2\%}
\InsertImageD{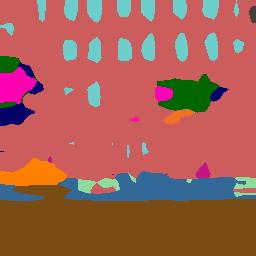}{76.2\%}{47.8\%}
\InsertImageD{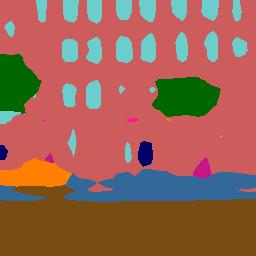}{78.9\%}{60.2\%}
\InsertImageS{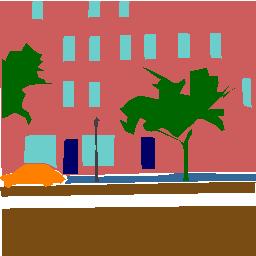}
\hspace{-1.75em}
\InsertImageS{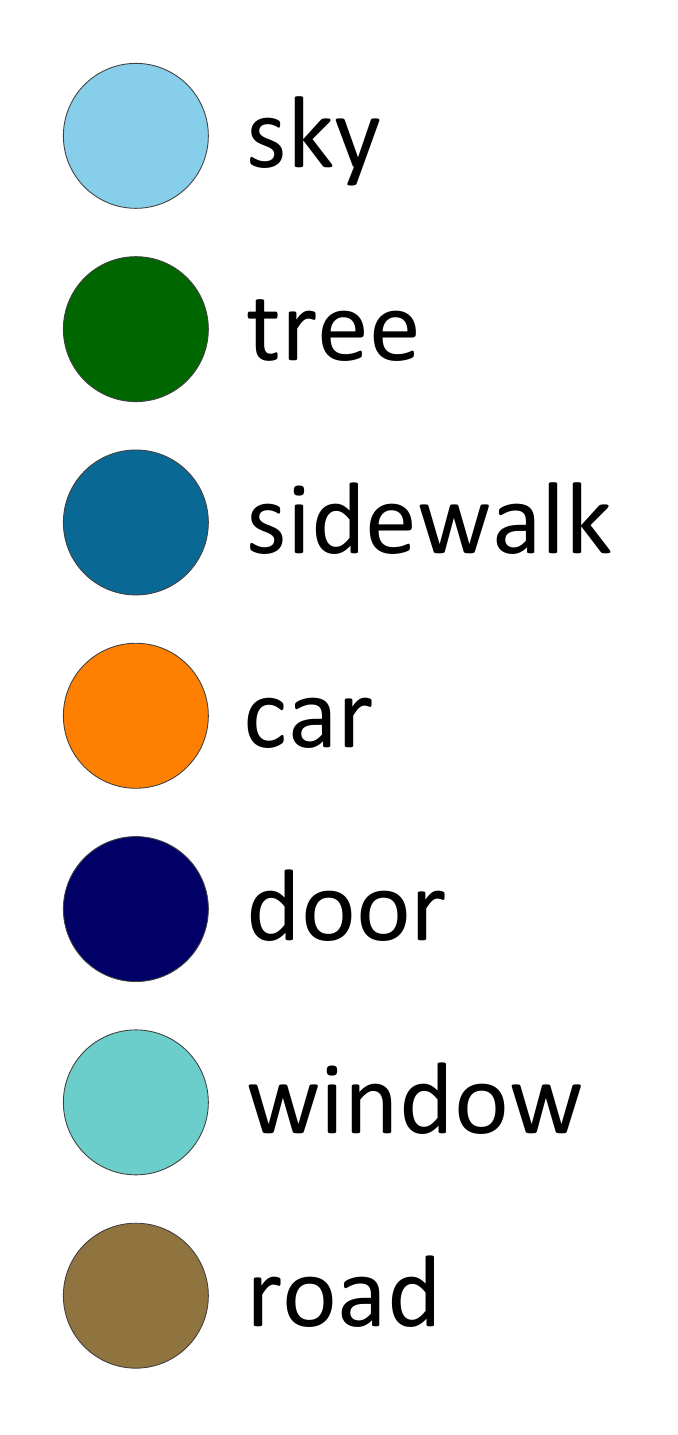}\\
\vspace{-0.25em}

\InsertImageS{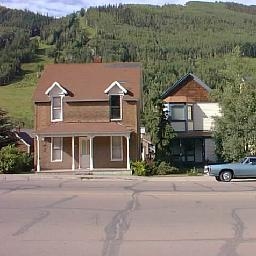}
\InsertImageS{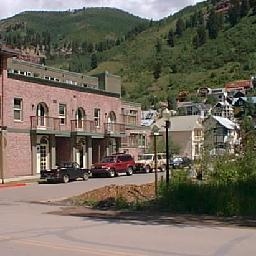}
\InsertImageS{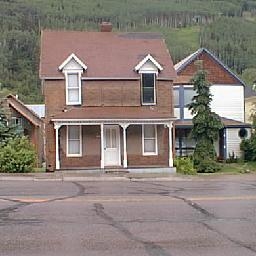}
\InsertImageS{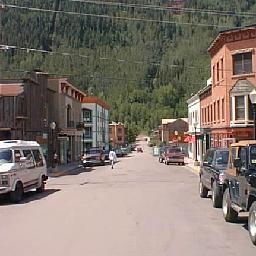}
\InsertImageD{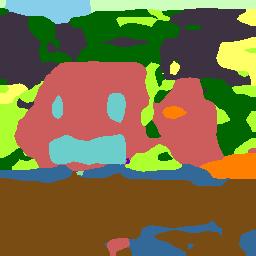}{66.8\%}{53.9\%}
\InsertImageD{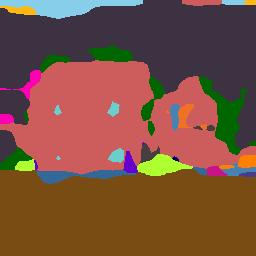}{79.9\%}{41.6\%}
\InsertImageD{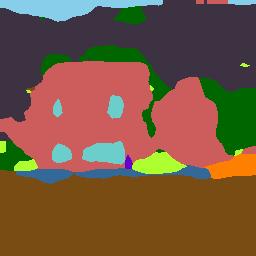}{81.8\%}{54.7\%}
\InsertImageS{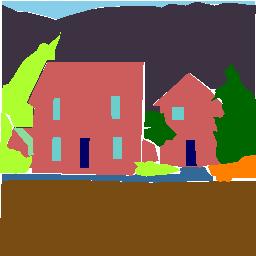}
\hspace{-1.75em}
\InsertImageS{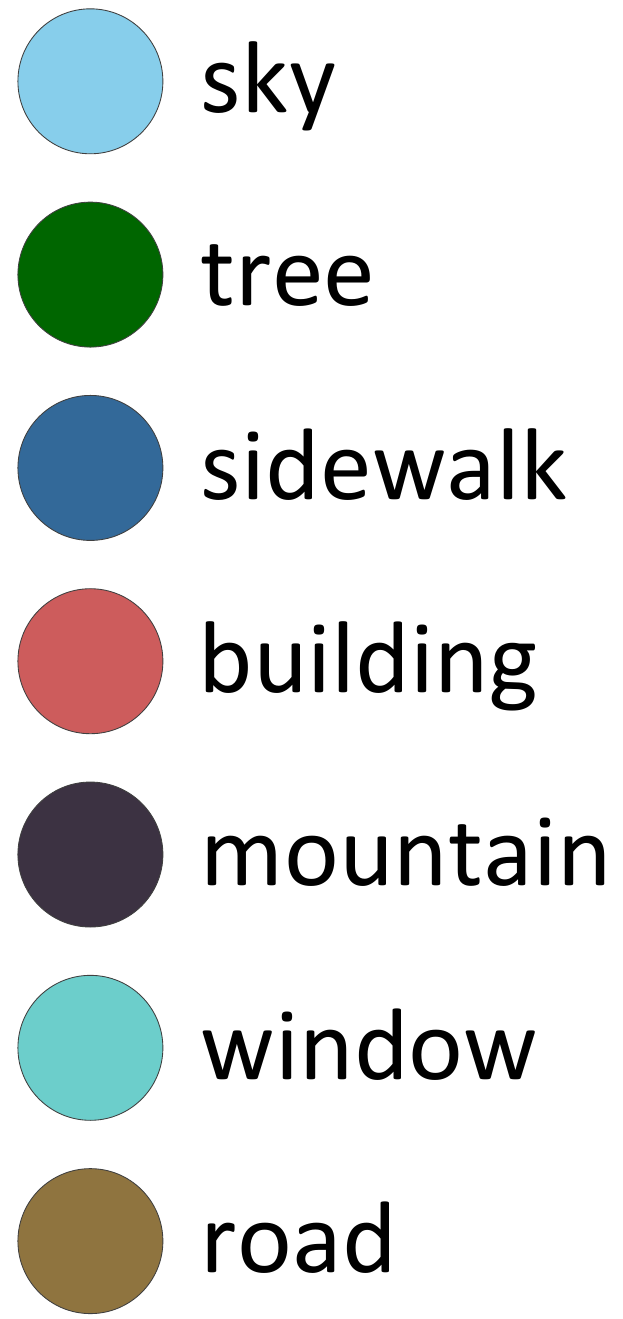}\\
\vspace{-0.25em}

\InsertImageS{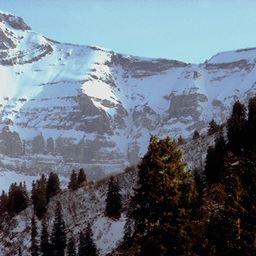}
\InsertImageS{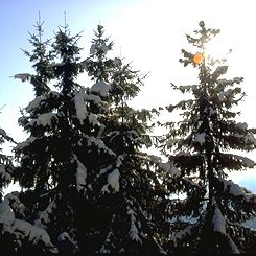}
\InsertImageS{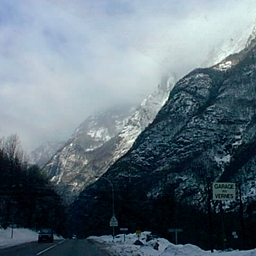}
\InsertImageS{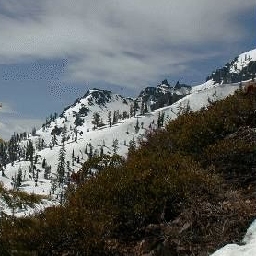}
\InsertImageD{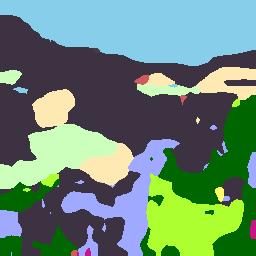}{65.1\%}{68.6\%}
\InsertImageD{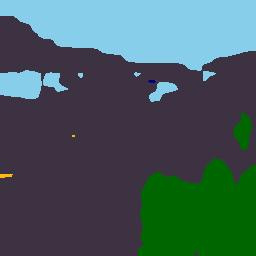}{84.2\%}{82.8\%}
\InsertImageD{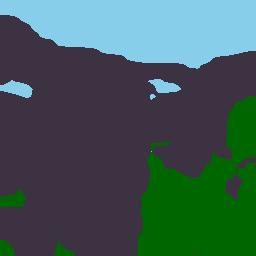}{88.6\%}{87.4\%}
\InsertImageS{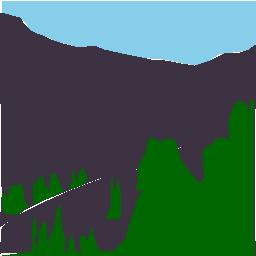}
\hspace{-1.75em}
\InsertImageS{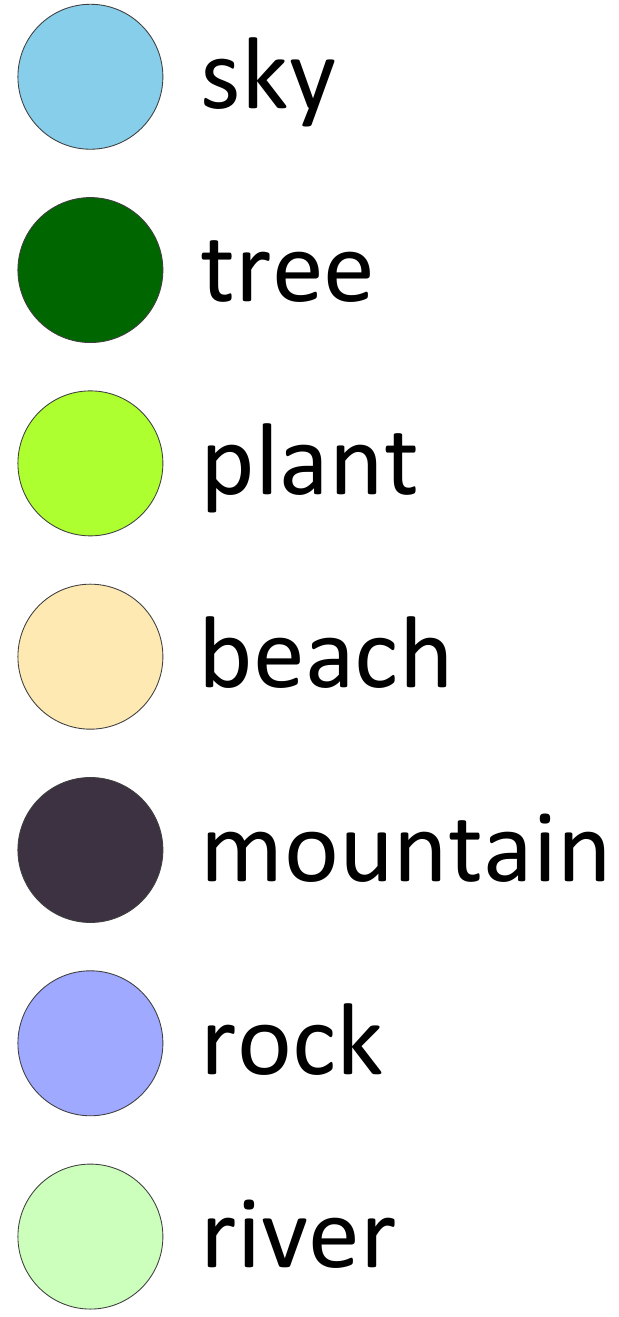}\\
\vspace{-0.25em}

\begin{picture}(1,1)
\setlength{\unitlength}{\textwidth}
\thicklines
\put(-0.505,0){\line(1,0){0.965}}
\end{picture}

\vspace{0.75em}
\InsertImageS{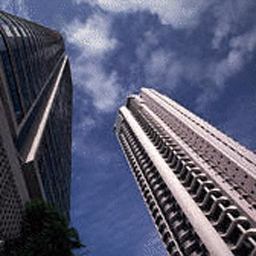}
\InsertImageS{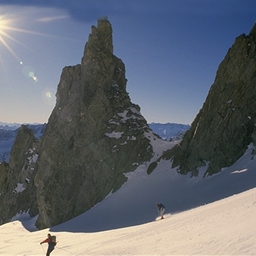}
\InsertImageS{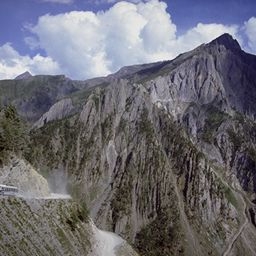}
\InsertImageS{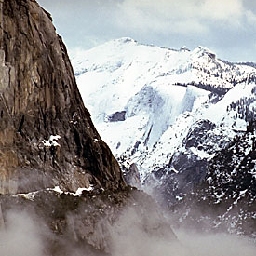}
\InsertImageD{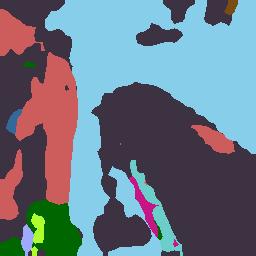}{50.9\%}{57.1\%}
\InsertImageD{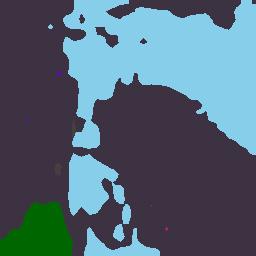}{36.2\%}{52.2\%}
\InsertImageD{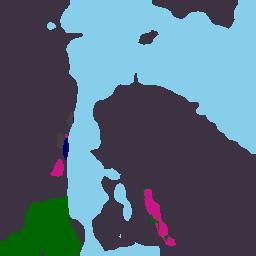}{39.1\%}{56.7\%}
\InsertImageS{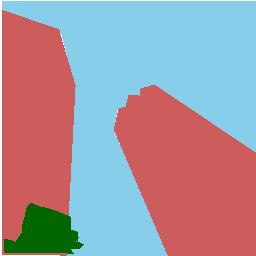}
\hspace{-1.75em}
\InsertImageS{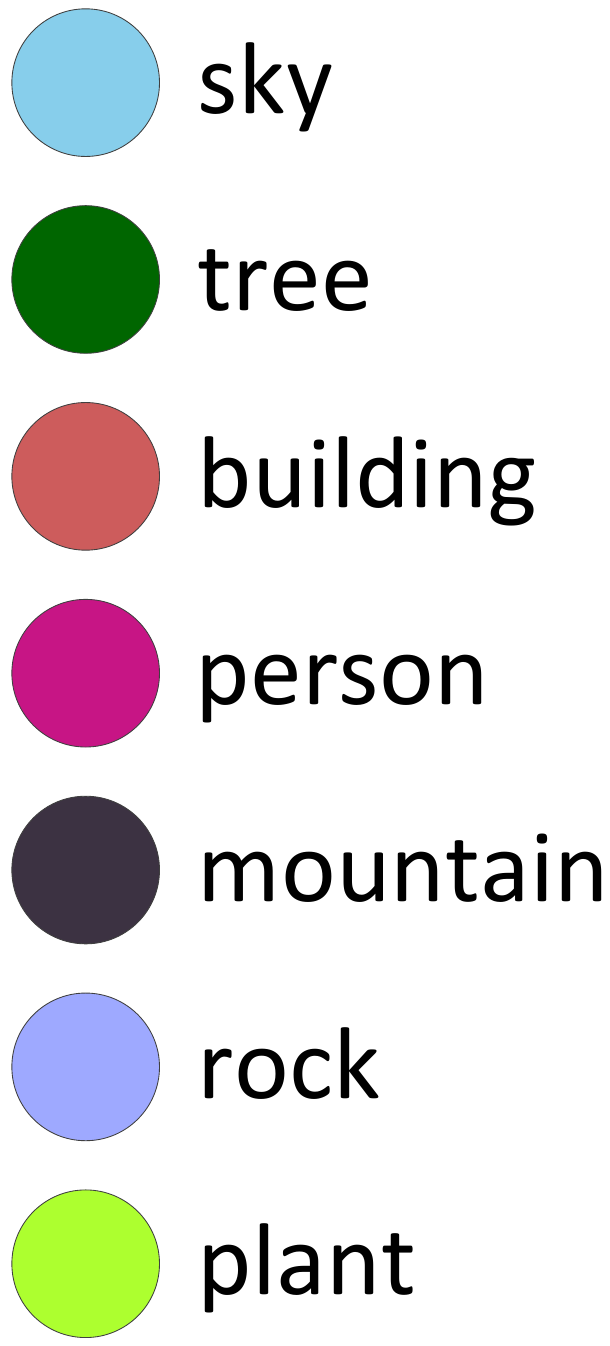}\\
\vspace{-0.5em}

\end{center}
\caption{Qualitative labeling examples on the SiftFlow dataset (best viewed in color). In each row, we show input images, top three nearest exemplar images, local labeling maps (output by the parametric model - CNN-Ensemble), global labeling maps (output by the non-parametric model), final label prediction maps (output by the integration model) and their ground truth maps respectively. The numbers outside and inside parentheses are global pixel accuracy (GPA) and average class accuracy (ACA) respectively.  The last row shows an example, where the \textit{local ambiguity} cannot be removed when their aggregated global features fail to reveal the true scene semantics. }
\label{Fig:qualitative_result}
\end{figure*}

The integration model applies Equation \ref{Equation:main} to integrate the local and global beliefs, which yields the un-normalized class likelihood of pixels. Based on which, the final prediction map is produced. Table \ref{Table:integration-result-siftFlow}, \ref{Table:integration-result-barcelona} list the quantitative performance on the SiftFlow and Barcelona dataset respectively.  As shown, our integration model is able to take advantage of both models, therefore it outputs more reliable label maps quantitatively and qualitatively. On one hand, \textit{it dramatically boosts the global pixel accuracy (GPA) compared to the local labeling (CNN-Ensemble)}: \textbf{\color{red}{5.9}\%} and \textbf{\color{red}{9.0}\%} GPA improvement on the SiftFlow and Barcelona dataset respectively. These results elucidate that a higher quality labeling prediction map can be obtained by enforcing a global scene constraint to the local classification. On the other hand, \textit{it also tremendously enhances the average class accuracy (ACA) compared to the global labeling}: \textbf{\color{red}{9.3}\%} and \textbf{\color{red}{4.1}\%} ACA improvement on the Siftflow and Barcelona dataset respectively. These results indicate that some object classes are ignorable in a global view (e.g. a small bird flies in broad sky), and the integration of local cues is akin to delineating objects in a global scene image. A number of qualitative examples are demonstrated in Figure \ref{Fig:qualitative_result}.

In comparison with other representation networks \cite{farabet2013learning} \cite{pinheiro2014recurrent}, our integration model outperforms them by a large margin. Thus, enforcing a global scene constraint to local classification is a promising solution to alleviate \textit{local ambiguities}. Furthermore, we compare our integration model with state-of-the-art counterparts. As listed in Table \ref{Table:integration-result-siftFlow}, our method achieves very competitive results on the SiftFlow dataset. \textcolor{black} {It's important to note that \cite{george2015image} uses 20 types of low-level features and further augments it with Fisher Vector (FV) \cite{perronnin2010large} descriptor (based on the SIFT feature) to represent each image superpixel. In contrast, we only use very compact (64-dimension) features learned from CNNs. In addition, \cite{long2015fully} adopts the very deep CNNs \cite{simonyan2014very} pretrained on the large-scale ImageNet dataset \cite{russakovsky2014imagenet} to generate the local features, whereas we utilize much shallower CNNs, which are trained with image patches only from the target dataset.}
Meanwhile, Table \ref{Table:integration-result-barcelona} clearly manifests that our method achieves the new best results on the more challenging Barcelona dataset, which significantly outperforms the previous state-of-the-arts.

\subsection{Analysis of Local and Global Beliefs}

\begin{table}
\begin{center}
\begin{tabular}{|l|c|cc|}
\hline
 & Frequency & CNN-Ensemble  & Integration Model \\
\hline
\hline
sky & 27.1\% & 93.5\%  & \textbf{96.7}\% \\
building & 20.2\% & 80.6\%  & \textbf{88.0}\% \\
tree & 12.6\% & 74.8\%  & \textbf{84.7}\% \\
mountain & 12.4\% & 71.5\%  & \textbf{80.3}\% \\
road & 6.93\% & 76.8\%  & \textbf{85.8}\% \\
sea & 5.6\% & 59.7\%  & \textbf{75.0}\% \\
field & 3.64\% & 33.1\%  & \textbf{37.6}\% \\
car & 1.6\% & 80.5\%  &78.7\% \\
sand & 1.41\% & 33.6\%  & \textbf{37.5}\% \\
river & 1.37\% & 44.0\%  & \textbf{50.1}\% \\
plant & 1.33\% & 27.5\%  & 7.56\% \\
grass & 1.22\% & 75.2\%  & 72.6\% \\
window & 1.07\% & 45.1\%  & 33.8\% \\
sidewalk & 0.89\% & 60.2\%  & 53.7\% \\
rock & 0.85\% & 23.8\%  & 13.0\% \\
door & 0.26\% & 36.8\%  & \textbf{40.2}\% \\
fence & 0.24\% & 44.5\%  & \textbf{44.6}\% \\
person & 0.23\% & 30.9\%  & \textbf{44.2}\% \\
staircase & 0.18\% & 45.4\%  & 24.6\% \\
awning & 0.11\% & 7.54\%  & \textbf{14.5}\% \\
sign & 0.11\% & 33.4\%  & \textbf{50.3}\% \\
boat & 0.06\% & 2.22\%  & \textbf{3.05}\% \\
crosswalk & 0.05\% & 85.4\%  & 74.1\% \\
bridge & 0.04\% & 18.8\%  & \textbf{20.4}\% \\
pole & 0.04\% & 2.43\% & \textbf{7.1}\% \\
balcony & 0.03\% & 34.4\%  & 29.9\% \\
bus & 0.03\% & 0.05\%  & 0.05\% \\
streetlight & 0.02\% & 15.1\%  & 2.36\% \\
sun & 0.01\% & 93.3\%  & 86.0\% \\
bird & $\approx 0$ & 13.2\% & \textbf{28.3}\% \\
\hline
\hline
GPA & - & 75.3\%  & \textbf{81.2}\% \\
ACA & - & 44.8\% & \textbf{45.5}\% \\
\hline
\end{tabular}
\end{center}
\caption{Per-class accuracy comparison of different models on the SiftFlow dataset. The statistics of class frequency is obtained in test images. }
\label{Table:siftFlow_class_accuracy}
\end{table}

In this section, we first compare the characteristics of local and global beliefs. By looking into the local/global labeling results quantitatively and qualitatively, we notice that the global labeling is more likely to output globally consistent label maps, while the local labeling focuses on discriminating local regions. Specifically, the global belief prioritizes the background classes, whereas the local belief works excellently on differentiating locally distinct object classes (especially small size classes). Hence, the global belief defines the global scene prior, while local belief preserves the locality information. Moreover, the reciprocity of local and global beliefs necessitates the competitive performance of our integration model. A glimpse of these properties are depicted in the qualitative labeling examples of Figure \ref{Fig:qualitative_result}. Meanwhile, It is interesting to see that the global labeling outperforms local labeling (CNN-GS), which illuminates the significance of global context for local classification.
\footnote{Under this situation, these two models are working on the same pixel features, which are generated by truncated CNN-GS.}

We further investigate the per-class accuracy changes of the parametric model after it integrates with the non-parametric model. Table \ref{Table:siftFlow_class_accuracy} shows the quantitative results on the SiftFlow dataset: the integration model boosts the accuracy significantly for frequent classes, while slightly washes away some rare ``object" classes. In detail, the global belief is more helpful for classes which are more stable in positions, and large-size classes are preferred because the target classes to be included in the nearest exemplars. Overall, our integration model is able to achieve very competitive average class accuracy (ACA), and in the same time dramatically improve the qualitative labeling results. As evidenced by Table \ref{Table:cnn-result-comp} and \ref{Table:integration-result-barcelona}, similar results are also observed on more challenging Barcelona datasets.


\section{Conclusion}
\label{Section:Conclusion}
In this paper, we first present a very effective parametric model - \textbf{CNN-Ensemble} - for local classification. The CNN components in the CNN-Ensemble are trained from image patches which are very different in the form of class frequency distribution. Therefore, each CNN component learns nonidentical visual patterns, and their decision fusion gives rise to more accurate local beliefs. Then, we alleviate the notorious \textit{local ambiguity} problem by introducing a global scene constraint, which is mathematically achieved by adding a global energy term to the labeling energy function, and it is practically estimated in a non-parametric framework. Furthermore, A large margin based CNN metric learning method is also proposed for better global belief estimation. The final class likelihood of pixels are obtained by integrating local and global cues. The outstanding quantitative and qualitative results on the challenging SiftFlow and Barcelona datasets illuminate the effectiveness of our methods.

\vspace{6pt}
\noindent
{\bf Acknowledgements:} This research was carried out at the Rapid-Rich Object Search (ROSE) Lab at the Nanyang Technological University, Singapore. The ROSE Lab is supported by the National Research Foundation, Singapore, under its Interactive Digital Media (IDM) Strategic Research Programme.
The research is also supported by Singapore Ministry of Education (MOE) Tier 2 ARC28/14, and Singapore A*STAR Science and Engineering Research Council PSF1321202099. The authors would also like to thank NVIDIA for their generous donation of GPU.


\bibliographystyle{IEEEtran}
\bibliography{tip-2015-shuai}

\end{document}